%% file: WileyMSPmain.tex
\documentclass[]{WileyMSP-template}

\usepackage{textcomp, gensymb}
\usepackage[inline]{enumitem}

\usepackage[locale=US, per-mode=reciprocal]{siunitx}

\usepackage{hyperref}

\usepackage{orcidlink}

\usepackage{tikz}
\usetikzlibrary{patterns}
\usetikzlibrary{positioning}
\usetikzlibrary{calc}
\usetikzlibrary{shapes.geometric}
\usetikzlibrary{shapes}

\usepackage{pgfplots}
\usepgfplotslibrary{fillbetween}
\usepgfplotslibrary{units}
\usepgfplotslibrary{statistics}
\pgfplotsset{compat=1.18}

\pgfplotsset{
	box plot/.style={
		/pgfplots/.cd,
		black,
		only marks,
		mark=-,
		mark size=1em,
		/pgfplots/error bars/.cd,
		y dir=plus,
		y explicit,
	},
	box plot box/.style={
		/pgfplots/error bars/draw error bar/.code 2 args={%
			\draw ##1 -- ++(1em,0pt) |- ##2 -- ++(-1em,0pt) |- ##1 -- cycle;
		},
		/pgfplots/table/.cd,
		y index=2,
		y error expr={\thisrowno{3}-\thisrowno{2}},
		/pgfplots/box plot
	},
	box plot top whisker/.style={
		/pgfplots/error bars/draw error bar/.code 2 args={%
			\pgfkeysgetvalue{/pgfplots/error bars/error mark}%
			{\pgfplotserrorbarsmark}%
			\pgfkeysgetvalue{/pgfplots/error bars/error mark options}%
			{\pgfplotserrorbarsmarkopts}%
			\path ##1 -- ##2;
		},
		/pgfplots/table/.cd,
		y index=4,
		y error expr={\thisrowno{2}-\thisrowno{4}},
		/pgfplots/box plot
	},
	box plot bottom whisker/.style={
		/pgfplots/error bars/draw error bar/.code 2 args={%
			\pgfkeysgetvalue{/pgfplots/error bars/error mark}%
			{\pgfplotserrorbarsmark}%
			\pgfkeysgetvalue{/pgfplots/error bars/error mark options}%
			{\pgfplotserrorbarsmarkopts}%
			\path ##1 -- ##2;
		},
		/pgfplots/table/.cd,
		y index=5,
		y error expr={\thisrowno{3}-\thisrowno{5}},
		/pgfplots/box plot
	},
	box plot median/.style={
		/pgfplots/box plot
	}
}

\usepackage{amsmath}

\usepackage{cleveref}

\usepackage[nolist]{acronym}
\begin{acronym}
	\acro{CPM Lab}{Cyber-Physical Mobility Lab}
	\acro{CPM Remote}{Cyber-Physical Mobility Lab Remote Access}
    \acro{CAV}{Connected and Automated Vehicle}
    \acro{MPC}{Model Predictive Control}
    \acro{NCS}{Networked Control Systems}
    \acro{VM}{Virtual Machine}
    \acro{UAV}{Unmanned Aerial Vehicle}
    \acro{ADAS}{Advanced Driver Assistance Systems}
    \acro{V2V}{Vehicle-to-Vehicle}
    \acro{V2I}{Vehicle-to-Infrastructure}
    \acro{V2X}{Vehicle-to-Everything}
    \acro{VRU}{Vulnerable Road User}
    \acro{FOV}{Field of View}
    \acro{GPS}{Global Positioning System}
    \acro{RTK}{Real-time Kinematic Positioning}
\end{acronym}

\newcommand{\associationmatrix}{{\pmb{A}}}
\newcommand{\costmatrix}{{\pmb{C}}}
\newcommand{\costmatrixthreshold}{{c_{max}}}

\newcommand{\objectlist}{{\mathcal{O}}}
\newcommand{\currentobjectlist}{{\mathcal{O}_{k}}}
\newcommand{\lastobjectlist}{{\mathcal{O}_{k-1}}}

\newcommand{\object}{{O}}

\newcommand{\boundingboxlist}{{\mathcal{B}}}
\newcommand{\boundingbox}{{B}}

\newcommand{\positionvector}{{\pmb{p}}}
\newcommand{\velocityvector}{{\pmb{v}}}
\newcommand{\positionx}{{p_x}}
\newcommand{\positiony}{{p_y}}
\newcommand{\positionz}{{p_z}}
\newcommand{\velocityx}{{v_x}}
\newcommand{\velocityy}{{v_y}}
\newcommand{\accelerationx}{{a_x}}
\newcommand{\accelerationy}{{a_y}}
\newcommand{\yaw}{{\theta}}
\newcommand{\yawrate}{{\omega}}
\newcommand{\yawraterate}{{\dot{\omega}}}
\newcommand{\length}{{l}}
\newcommand{\width}{{w}}
\newcommand{\height}{{h}}

\newcommand{\covariance}{{\pmb{P}}}
\newcommand{\variance}{{\sigma}}

\newcommand{\state}{{\pmb{x}}}
\newcommand{\stateinput}{{\pmb{u}}}
\newcommand{\measurement}{{\pmb{z}}}

\newcommand{\estimatedstate}{{\hat{\state}}}
\newcommand{\statecovariance}{{\covariance}_{\estimatedstate}}
\newcommand{\dimension}{{\pmb{d}}}
\newcommand{\dimensionelement}{{d}}

\newcommand{\estimateddimension}{{\hat{\dimension}}}
\newcommand{\estimateddimensionelement}{{\hat{\dimensionelement}}}

\newcommand{\measureddimension}{{\tilde{\dimension}}}
\newcommand{\dimensionvariance}{{\covariance}_{\estimateddimension}}
\newcommand{\measureddimensioncovariance}{{\covariance}_{\measureddimension}}

\newcommand{\measureddimensionelementi}{{\tilde{\dimensionelement}^i}}

\newcommand{\measureddimensionelementivariance}{{\variance}_{\measureddimensionelementi}}

\newcommand{\probability}{{p}}

\newcommand{\existenceprobability}{{\probability_e}}
\newcommand{\classificationvector}{{\pmb{c}}}
\newcommand{\classification}{{c}}

\newcommand{\statetransitionmatrix}{{\pmb{F}}}
\newcommand{\stateinputmatrix}{{\pmb{B}}}
\newcommand{\processnoisematrix}{{\pmb{Q}}}
\newcommand{\measurementmatrix}{{\pmb{H}}}
\newcommand{\transformationmatrix}{{\pmb{T}}}

\newcommand{\pointcloud}{{\mathcal{P}}}
\newcommand{\currentpointcloud}{{\mathcal{P}_{k}}}

\newcommand{\cluster}{{\rho}}

\newcommand{\objectpointcloud}{{\mathcal{P}_{\mathcal{O}}}}

\newcommand{\approxobjectpointcloud}{{\tilde{\mathcal{P}}_{\mathcal{O}}}}
\newcommand{\currentapproxobjectpointcloud}{{\tilde{\mathcal{P}}_{\mathcal{O},{k}}}}

\begin{document}

\pagestyle{fancy}

\title{Lidar-based Tracking of Traffic Participants with Sensor Nodes in Existing Urban Infrastructure}

\maketitle


\author{Simon Schäfer*}
\author{Bassam Alrifaee}
\author{Ehsan Hashemi}



\begin{affiliations}
S. Schäfer* \orcidlink{0000-0002-6482-2383}\\
Chair of Embedded Software, RWTH Aachen University, Aachen, Germany\\
schaefer@embedded.rwth-aachen.de
\vspace{0.2cm}

Prof. B. Alrifaee \orcidlink{0000-0002-5982-021X}\\
Department of Aerospace Engineering, University of the Bundeswehr Munich, Munich, Germany\\
bassam.alrifaee@unibw.de
\vspace{0.2cm}

Prof. E. Hashemi \orcidlink{0000-0002-6236-7516}\\
Department of Mechanical Engineering, University of Alberta, Edmonton, Canada\\
ehashemi@ualberta.ca
\end{affiliations}



\begin{abstract}
This paper presents a lidar-only state estimation and tracking framework, along with a roadside sensing unit for integration with existing urban infrastructure. Urban deployments demand scalable, real-time tracking solutions, yet traditional remote sensing remains costly and computationally intensive, especially under perceptually degraded conditions. Our sensor node couples a single lidar with an edge computing unit and runs a computationally efficient, GPU-free observer that simultaneously estimates object state, class, dimensions, and existence probability. The pipeline performs: (i) state updates via an extended Kalman filter, (ii) dimension estimation using a 1D grid-map/Bayesian update, (iii) class updates via a lookup table driven by the most probable footprint, and (iv) existence estimation from track age and bounding-box consistency. Experiments in dynamic urban-like scenes with diverse traffic participants demonstrate real-time performance and high precision: The complete end-to-end pipeline finishes within \SI{100}{\milli\second} for \SI{99.88}{\%} of messages, with an excellent detection rate. Robustness is further confirmed under simulated wind and sensor vibration. These results indicate that reliable, real-time roadside tracking is feasible on CPU-only edge hardware, enabling scalable, privacy-friendly deployments within existing city infrastructure. The framework integrates with existing poles, traffic lights, and buildings, reducing deployment costs and simplifying large-scale urban rollouts and maintenance efforts.
\end{abstract}


\section*{Open Material}
\label{sec:openmaterial}

\begin{flushleft}
\textbf{Code} \href{tbd}{Published on acceptance} \\
\end{flushleft}

\section{Introduction}
\label{sec:intro}

Urban traffic management is increasingly relying on remote sensing technologies like lidar to improve both safety and efficiency.
These systems can offer significant benefits, such as providing real-time warning systems at intersections or enhancing traffic flow monitoring.
However, traditional remote sensing solutions are often costly, making them impractical for large-scale deployment.
There's a need for scalable, cost-effective solutions that can integrate with existing infrastructure and provide reliable data to improve urban mobility management.

Sensor nodes embedded in urban infrastructure, such as light poles, traffic signals, or buildings, present a promising solution to these challenges.
By using infrastructure already in place, these nodes enable the collection of data without the need for extensive new construction.
These sensor nodes can reduce deployment costs and offer a more affordable alternative to traditional sensing technologies.
However, using existing infrastructure also poses challenges, such as the potential for inadequate sensor placement due to limited available mounting locations or restrictions on mounting height.

Our proposed solution focuses on a state estimation framework that utilizes a lidar-only sensor node, equipped with a robust state observer, designed to track traffic participants in real time.
By using edge computing for processing, the system can handle real-time data without relying on cloud resources or expensive computational power.
This minimalistic setup is ideal for urban environments where computational efficiency is key.

Multimodal sensor information processing has been widely studied to improve state estimation accuracy, especially through the fusion of data from multiple sensor modalities such as cameras, lidar, and radar.
However, multimodal fusion introduces significant challenges, including data synchronization and processing complexity.
In contrast, our approach simplifies the system architecture by using a single lidar sensor for state estimation while still ensuring robust tracking.
Additionally, the system is designed to maintain privacy compliance by not transmitting raw sensor data, thus ensuring sensitive information remains secure while also optimizing operational efficiency.

We evaluate our proposed system using a dataset collected at the University of Alberta (UofA), demonstrating promising results.
Our system achieved a tracking accuracy of around \SI{0.5}{\meter}, detecting $100\%$ of the objects in the scene.
Additionally, we tested the system under simulated wind conditions, with sensor shaking.
These results showed a marginal decrease in tracking accuracy, indicating the robustness and stability of the observer in dynamic conditions.
As part of our open materials, we also provide MATLAB examples for some of the key algorithms in the observer, contributing to further research and development in this area.

The remainder of the paper is organized as follows: \Cref{sec:related_work} presents the results of our literature research about similar systems.
\Cref{sec:sensor_node} provides an overview of the proposed sensor node design.
\Cref{sec:observer} presents the state observer used for real-time tracking of traffic participants.
\Cref{sec:results} details the evaluation of the state observer’s performance, while \Cref{sec:conclusion} concludes the paper with a summary of the findings and discusses potential directions for future work.

\section{Related Work}
\label{sec:related_work}

\subsection{Infrastructure Lidar Deployments and Surveys} 

Infrastructure-based lidar for traffic monitoring has often been realized with multiple sensors, typically arranged around intersections for broad coverage. An early representative example is \cite{zhao2012detection}, where the authors synchronize four lidars to monitor an entire intersection, detecting bounding boxes of traffic participants. While they provide position and orientation estimates, their system omits classification and does not run in real time, relying instead on an offline rule-based association scheme. The authors of \cite{kloeker2020highprecision} similarly deploy four lidar sensors but use deep learning (PointPillars \cite{lang2019pointpillars}) to localize vehicles, returning high-precision 3D bounding boxes and classes. Such multi-lidar solutions reduce occlusion issues but are more expensive and complex to install and usually require additional infrastructure or changes to existing infrastrcuture.

Single-lidar approaches are simpler but can struggle with blind spots and partially visible objects. The authors of \cite{zhao2021lshape, song2023lshape} propose a L-shape-based detection criterion specifically for vehicles where only one edge or side of a vehicle is visible. They also discuss the algorithm’s real-time feasibility yet do not detail the underlying hardware platform. In a different single-lidar design, Bai \emph{et al.}~\cite{bai2023cyber} track road users with a deep neural network on a dedicated edge server, achieving similar localization error to our approach. However, their system processes the full 3D point cloud and regularly exceeds the real-time deadline.

Several broader surveys highlight that few production-grade intelligent transportation systems (ITS) rely on lidar alone. Gress \emph{et al.}\cite{gress2024intelligent} review 53 ITS applications, noting that only 12 involved lidars at all, and none used lidar as a standalone tracking source. Pandharipande \emph{et al.}\cite{pandharipande2023sensing} similarly observe that machine learning pipelines currently dominate lidar usage in automated driving, but such pipelines require high-end hardware with considerable power draw and GPU acceleration.

\subsection{Approaches to Lidar Data Processing: Model-Based vs. Learning-Based}

Existing roadside tracking pipelines generally fall into two categories: model-based methods that apply geometric or rule-based logic, and neural networks that segment the point cloud or detect bounding boxes directly from raw point clouds. The model-based branch often relies on clustering, bounding-box fitting, and a state-estimation filter such as a Kalman or particle filter. For example, \cite{zhao2019detection} remove static objects via a 3D background filter, adapt clustering thresholds based on distance, and use a small neural network for basic object type discrimination (vehicles vs. pedestrians), followed by a Kalman filter for trajectory tracking. Those approaches are feasible es shown by \cite{liu2023gnnpmb} who systematically compare machine learning approaches to model-based pipelines on the nuScenes dataset \cite{caesar2020nuscenes} and find that simpler, model-based methods can match or exceed advanced neural networks in some scenarios. Additionally \cite{lee2022moving} shows that a particle filter outperforms simpler Kalman filter in complex, dense traffic but is measurably slower, important if the method is suppose to be executed on embedded hardware with limited computational overhead.

Machine-learning approaches concentrate on 3D point-cloud segmentation or 3D bounding-box detection. Mai \emph{et al.}\cite{mai2020semantic} summarize a variety of pointwise semantic-segmentation algorithms, noting that many exceed \SI{200}{\milli\second} per frame even on powerful graphics processors. Additionally, a lot of those methods are trained from an vehicle perspective rather than an infrastrucuter perspective. Zhou \emph{et al.}\cite{zhou2022leveraging} propose transferring those pre-trained vehicle detectors to a roadside viewpoint. While promising their adaptation step lowers accuracy compared to the original domain. Research into purely learning-based pipelines that incorporate tracking e.g., \cite{loc2025accurate} can approach very fast inference on a GPU, but combined power usage of those systems often surpasses typical roadside budgets. Eventhough they do create the most accurate results as for exmaple shown by \cite{li2023high} who explore point-cloud registration plus clustering, then project data onto images for CNN-based classification, achieving around \SI{0.09}{\meter} position error but requiring up to \SI{500}{\milli\second} per frame for processing. This gap is tackled by other authors e.g., \cite{xia2023alightweight} minimize inference time of popular networks down to \SI{14}{\milli\second} but the requirement for specialized GPU hardware remains. 

The work \cite{zhou2018voxelnet} proposed VoxelNet, an end-to-end object detection approach. Later, other authors extended VoxelNet to VoxelNeXt in \cite{chen2023voxelnext}. Both approaches perform 2D bounding box extraction on raw point clouds and are evaluated on similar GPUs with inference times of \SI{32}{\milli\second} and \SI{62}{\milli\second}, respectively.
Authors of \cite{barrera2021birdnetplus} present an end-to-end detection framework that performs 3D bounding box extraction comparable to our detection approach. They evaluated the approach on a full-size GPU and achieved inference times of \SI{52}{\milli\second} for the detection step. However, the approach does not contain a tracking component.
Similar to those approaches, YOLO3D \cite{ali2018yolo3d} also performed well in detecting 3D oriented bounding boxes. In contrast to other authors, they analyzed the inference times with different point cloud resolutions. The results showed that the time ranged from \SI{13.5}{\milli\second} to \SI{41.25}{\milli\second}.

Regardless of their basis, many learning-centered methods we encountered during our literature review skip essential tracker components such as existence probability estimation or multi-target association or focus on other parts of the perception pipeline. For example the method in \cite{cheng20233dvehicle} only handles vehicle bounding boxes, ignoring pedestrians or cyclists; Fang \emph{et al.}\cite{fang20213dsiamrpn}  concentrates on single-object tracking. Others methodes fuse lidar with cameras or radar, as in \cite{borba2023increasing, han2023flexsense}, making sensor deployment more complex and reliant on good lighting conditions. Meanwhile, power consumption remains a pressing concern: Smal \emph{et al.}\cite{smal2022taksdriven} attempt to reduce power draw by adaptively adjusting the lidar frame rates, and \cite{latotzke2023fpga} port the preprocessor to FPGAs, cutting power needs by 42\%.

\subsection{Observations and Our Contribution} Most existing infrastructure-based solutions either use multiple lidars for comprehensive intersection coverage or adopt a single lidar with machine learning for increased detection accuracy. However, \cite{zhao2021lshape} and \cite{song2023lshape} show, rule-based bounding-box fitting can be effective for vehicle detection, especially if the sensor overlooks them from above. Their works, along with  \cite{wu2024realtime} and \cite{lin2023identification}, highlight that prefiltering static elements and exploiting shape-based logic is a feasible approach. However, the majority of works omits existence probability estimates, classification, or explicit total runtime analysis. Our work addresses these gaps by presenting the complete end-to-end pipeline, beginning with designing a standalone lidar-only sensor node that incorporates a point cloud preprocessing pipeline for static object removal, bounding box detection through clustering, tracking objects using a model-based Kalman filter, and finally classifying them based on their estimated bounding box, all executed on an embedded CPU. This approach, being inherently deterministic and predictable, offers an explainable and safe solution compared to the more prevalent machine learning approaches. Moreover, it avoids the substantial power demands of GPUs.
Additionally, we dedicate attention to an explicit analysis of the real-time applicability of our approach.

\section{Sensor Node}
\label{sec:sensor_node}

Our sensor node integrates with existing urban infrastructure, such as light poles, traffic lights, and buildings, enabling quick and cost-effective deployment without new construction.
This approach minimizes disruption to urban planning and reduces installation effort by eliminating the need for dedicated sensor towers.
However, this approach also faces certain challenges, such as the possibility of inadequate sensor placement, limited number of possible mounting locations, and height restrictions on existing infrastructure.
Additionally, access to a reliable power supply may be unavailable in certain areas, requiring compatibility with an off-grid powering solutions such as solar panels or batteries.
Despite these challenges, the integration with existing urban infrastructure provides a scalable solution for large-scale deployments of lidar-based sensing systems in cities.

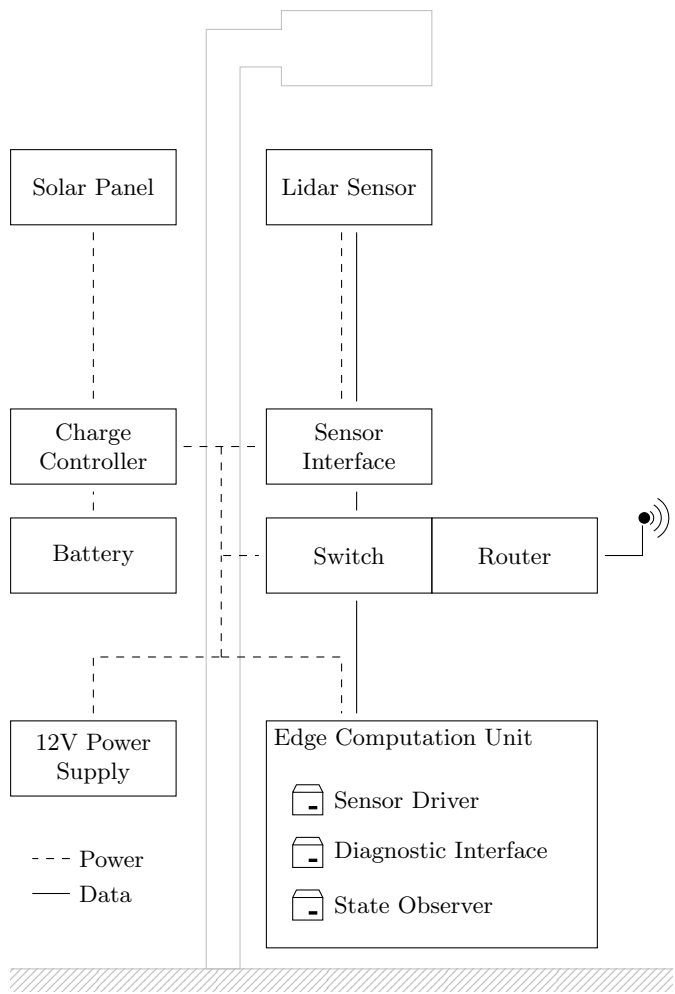
\begin{figure}[tbp]
    \centering
    \input{construction}
    \caption{Schematic showing the sensor node construction, including data and power connections between its components and the containerized software running on the computation unit.}
    \label{fig:overview_software}
\end{figure}

The sensor node employs a split design, as illustrated by \cref{fig:overview_software}, placing the lidar-based sensing unit on the pole and the edge computation unit and battery storage at ground level. This arrangement reduces the weight carried by the pole and simplifies maintenance tasks, an approach already commonplace in traffic-light installations. A directional lidar provides denser point clouds in the region of interest compared to traditional 360° models, particularly when pitched downward from a pole.

A T-shaped mounting arm, featuring a custom bracket design, attaches to various pole types. Traditional strap-style mounts prove unsuitable for the node’s weight, prompting the use of clamp mounts instead. The lidar itself rests beneath the outward-extending T-arm, leaving space for potential upgrades, such as cameras or additional sensors. In keeping with this modular concept, the system uses an NVIDIA Jetson Orin NX as the edge computation unit. Although the present solution runs solely on CPU resources, future development, including the addition of cameras, may use GPU-accelerated algorithms. As evidenced in \Cref{sec:results_runtime}, a less powerful CPU-only unit could still suffice, given that the system currently utilizes only 40\% of the Jetson's CPU.

Our architecture aims for minimal manual oversight. It comprises four main components, a sensor, an edge computation unit, an Ethernet switch, and a cellular router, connected via a local network. Each element initiates the necessary services automatically. To accommodate varied installation scenarios, the system accepts either mains power or a 12V battery. In the event of off-grid deployment, an on-board \SI{100}{\ampere \hour} battery and two \SI{200}{\watt} solar panels support continuous \SI{24}{\hour} operation in typical spring and summer conditions at the University of Alberta. At peak loads, the node draws approximately \SI{100}{\watt}, providing up to \SI{12}{\hour} of autonomous runtime before requiring recharge.

On the software side, the edge computation unit runs three key modules: the sensor driver, state observer, and a diagnostic interface, each packaged in Docker containers for over-the-air updates and automatic restarts on failure. A remotely accessible web interface allows the monitoring of key parameters such as update rates and sensor operating temperatures. The entire system follows a service-oriented architecture using ROS2 \cite{macenskiROS2022}, simplifying future component additions or removals. Notably, the system is privacy compliant by transmitting no raw sensor data, the pipeline processes information locally using the methods in \Cref{sec:observer} and then shares object lists over the network. This approach safeguards sensitive data and eases bandwidth requirements.


\section{Notation}
\label{sec:notation}

In this paper, we adopt the following notation conventions: 

\begin{itemize}
    \item Lowercase bold letters, such as $\pmb{a}$, represent vectors, while uppercase bold letters, such as $\pmb{A}$, denote matrices.
Calligraphic letter, like $\mathcal{A}$, donate sets of elements.
    \item The superscript $\left(.\right)^T$ indicates the transpose of a vector or matrix, and $\left(.\right)^{-1}$ represents the inverse of a matrix.
An estimate of the true value $\left(.\right)$ is denoted by $\hat{\left(.\right)}$.
    \item The subscript $\left(.\right)_k$ specifies the value of $\left(.\right)$ at time step $k$, and $\left(.\right)_{j,k}$ refers to the $j$-th value of $\left(.\right)$ at time step $k$.
Additionally, $\left(.\right)^i$ designates the $i$-th component of a selected set, where $i \in \{a, b, c\}$.
\end{itemize}

\section{Lidar-based State Observer Design}
\label{sec:observer}

Object tracking and detection in this paper rely exclusively on rule-based lidar processing.
This approach is computationally efficient and can be executed on a CPU, minimizing the computational load on the edge node.
This efficiency makes it well-suited for real-time applications with limited processing power.

The following subsections detail the steps involved in the proposed method, illustrated in \cref{fig:discretization_dimension}:

\begin{enumerate}
    \item Preprocessing of the point cloud: Initial filtering and transformation to prepare the raw lidar data.
    \item Object detection in the preprocessed point cloud: Identification of potential objects based on spatial clustering and geometry analysis.
    \item Association of detected objects: Linking new detections with already detected objects from the last timestep.
    \item State fusion of objects: Integration of detection and tracking data using an adaptive Kalman filter approach to estimate object states.
    \item Dimension estimation of objects: Application of a grid map approach with binary Bayes filters to estimate object dimensions.
    \item Existence probability estimation: A heuristic-based existence estimator.
    \item Classification of objects: Probabilistic estimation of object types based on dimension estimates.
    \item Object management: Handling the lifecycle of objects, including initialization, and removal.
\end{enumerate}

\begin{figure}[tbp]
    \centering
    \input{overview}
    \caption{Overview of the data flow in the proposed lidar-based object tracker.
The approach is divided into two main applications: the lidar preprocessor (top) and the object tracker (bottom).
Rectangles represent processes, while rounded rectangles indicate model definitions.}
    \label{fig:overview}
\end{figure}
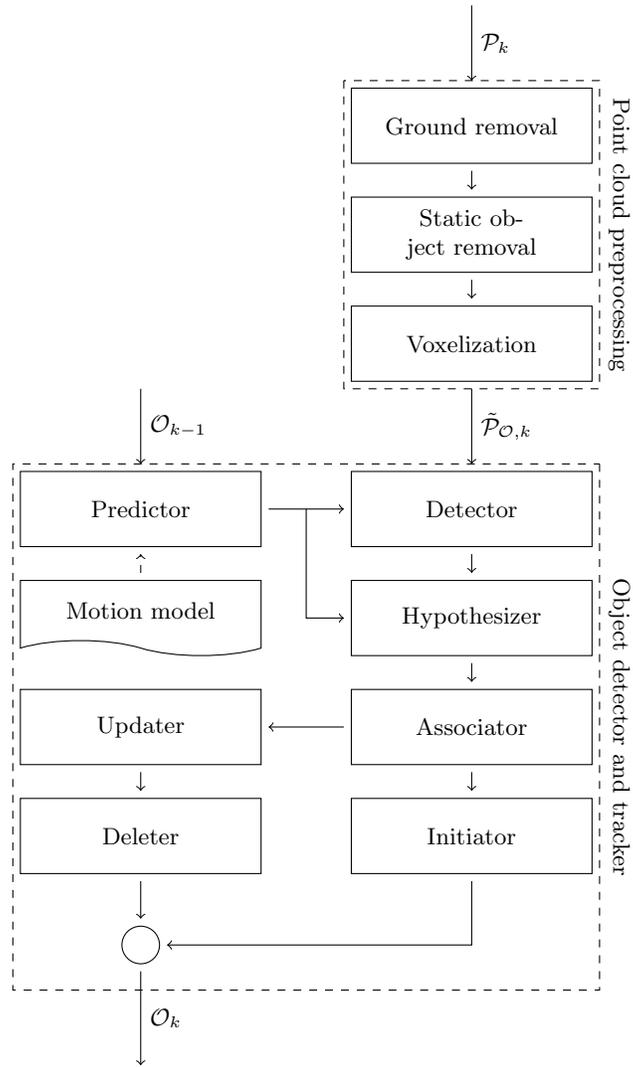

\subsection{Definitions}
\label{sec:definitions}

This paper begins with the preprocessing of point clouds, denoted as $\pointcloud$, collected by the sensor node.
A point cloud is a set of points in three-dimensional space, represented as
\begin{equation*}
    \pointcloud = \{(\positionx, \positiony, \positionz, i)_1, \dots, (\positionx, \positiony, \positionz, i)_M\}.
\end{equation*}
Each point comprises spatial coordinates $\positionx, \positiony, \positionz$ and an intensity value $i$, which measures the return signal strength from the sensor.
This type of data is commonly referred to as an "XYZI" point cloud.
The inclusion of intensity values alongside spatial coordinates can provide additional information about the reflective properties of surfaces, aiding in the differentiation of materials or textures in the scene.
However, the approach presented in this paper does not require the use of intensity information, ensuring broader applicability even when intensity data is unavailable or unreliable.

The objective of the estimation algorithm is to create a global object list, denoted as
\begin{equation*}
    \objectlist_{G} = \{\object_1, \object_2, \ldots, \object_N\}, 
\end{equation*}
which consists of $N$ objects.
Each object, represented as $O_i$, is defined as 
\begin{equation*}
    O_i = \left[\estimatedstate, \statecovariance, \estimateddimension, \dimensionvariance, \existenceprobability, \classificationvector\right].
\end{equation*}
Here, $\estimatedstate$ is the estimated state, $\statecovariance$ is the state covariance matrix, $\estimateddimension$ is the estimated dimension, $\dimensionvariance$ is the dimension covariance matrix, $\existenceprobability$ is the existence probability, and $\classificationvector$ is the classification vector.
The state vector, $\state$, encapsulates the position and dynamic motion data as
\begin{equation*}
    \state = \left[\positionvector, \velocityvector, \yaw, \yawrate\right]^T,
\end{equation*} where $\positionvector = \left[\positionx, \positiony\right]^T$ is the position on the $x$- and $y$-axes, $\velocityvector = \left[\velocityx, \velocityy\right]^T$ is the velocity on the $x$- and $y$-axes, $\yaw$ is the orientation of the object, and $\dot{\yaw} = \yawrate$ represents the rate of change of the orientation.

The dimension vector, $\dimension = \left[\length, \width, \height\right]^T$, quantifies the dimensions of the rectangular cuboid bounding box $\boundingbox$, with components $\length$, $\width$, and $\height$ representing its length, width, and height, respectively.
The bounding box’s center corresponds to the position vector $\positionvector$, its orientation is given by the yaw angle $\yaw$, and it assumes zero roll and pitch relative to the zero-elevation coordinate system.

To handle multiple frames of reference, the following coordinate systems are defined:
(1) a global geographic coordinate system for georeferencing results
(2) a local universal transverse mercator (UTM) coordinate system with an origin near the sensor node, (3) a sensor node base link coordinate system at elevation zero below the sensor node, denoted by $\left(.\right)_{b}$, and
(4) a sensor node link oriented with the sensor at its mounting elevation, denoted by $\left(.\right)_{s}$.
The transformation matrix from the sensor link to the base link, $T_{s \rightarrow b}$, is defined as $T_{s \rightarrow b} = S_z \cdot R_{pitch}$, where $S_z$ shifts the point cloud to elevation zero and $R_{pitch}$ accounts for the sensor's downward pitch angle.
The point cloud $\pointcloud_s$ captured by the lidar sensor in the sensor link coordinate system is transformed into the base link coordinate system using the relation $\pointcloud_b = T{s \rightarrow b} \cdot \pointcloud_s$.
Object detection and tracking are performed in the base link coordinate system, and the resulting tracks are transformed into the local map link and stored in global geographic coordinates for downstream processing.

\subsection{Point Cloud Preprocessing}
\label{sec:point_cloud_preprocessing}

This section describes the preprocessing pipeline, whose primary objective is to reduce the number of points in the point cloud while retaining the necessary data to detect dynamic objects in the scene.
Although we evaluate the proposed method on a single testing site, we designed the algorithm to remain abstract and adaptable for deployment at other locations in the future.

~\\
The proposed object tracking algorithm functions within a 2.5D coordinate system set at elevation zero.
While objects maintain their three-dimensional bounding boxes in this system, their positional and orientation coordinates are limited to two dimensions.
This assumption necessitates transforming the point cloud into an elevation-zero coordinate system, which can either be the base or the local coordinate system (see \cref{sec:definitions}).

Since there is no inherent advantage to choosing one system over the other, and a unified reference frame for processing is necessary, we select the base coordinate system as our reference frame.
The transformation $T_{s \rightarrow b}$ facilitates the conversion of the point cloud from the sensor’s local coordinate system to the base coordinate system as follows:
\begin{equation}
    \pointcloud_b = T_{s \rightarrow b} \cdot \pointcloud_s,
\end{equation} 
where $\pointcloud_s$ represents the raw point cloud gathered by the sensor node, and $\pointcloud_b$ the same point cloud expressed in the base coordinate system.
This transformation establishes the coordinate system used for all subsequent steps.

To simplify notation and improve readability, we assume that all quantities in the following sections are defined in the base coordinate system unless explicitly noted.
As such, we omit the subscript $ (.)_b $ throughout the paper.

The preprocessing pipeline benefits from the sensor node’s stationary nature.
Unlike lidars mounted on moving platforms, this setup simplifies the removal of ground points, as the sensor node does not roll, pitch, or yaw during operation.
We filter ground points using a straightforward axis-based method that removes points below a predefined elevation threshold.
While this approach is effective in environments with relatively consistent elevation, the approach could be extend by using an elevation map to handle locations with significant elevation changes.
The elevation map would be part of the transformation $\transformationmatrix{s \rightarrow b}$ accounting for varying terrain.
However, as detailed in section \cref{sec:results}, the test site used in this paper does not exhibit substantial elevation variations, making the more straightforward axis-based approach sufficient for our purposes.

We express the ground removal process as:
\begin{equation}
    \pointcloud_{\mathcal{G}} = \{(\positionx, \positiony, \positionz)_j | \positionz \leq p_{z,min} \forall j \in \mathcal{P}\},
\end{equation}
where $ p_{z,\text{min}} $ represents the elevation threshold below which we classify as ground and subsequently remove.

A second advantage of a stationary sensor node is quickly identifying and removing static objects from the point cloud.
Static objects, by definition, do not move and generally do not contribute significant information for understanding the dynamic aspects of a scene.
While automating the identification of static objects is theoretically possible, the limited number of such objects in each scene allows us to manually define their locations and dimensions.
We classify these predefined static objects using 3D bounding boxes and remove them from the point cloud using a crop-box filter.
We express the crop-box filter as:
\begin{equation}
    \pointcloud_{\mathcal{S}} = \{(\positionx, \positiony, \positionz)_j | j \in \mathcal{S} \forall j \in \pointcloud \setminus \pointcloud_{\mathcal{G}} \},
\end{equation}
where $ \mathcal{S} $ represents the set of predefined static objects.

After removing both the ground and static objects, the remaining points constitute the object point cloud, expressed as:
\begin{equation}
    \objectpointcloud = \{(\positionx, \positiony, \positionz)_j | \left(\positionz > p_{z,min} \land j \notin \mathcal{S} \right) \forall j \in \mathcal{P}\}.
\end{equation}

The preprocessing pipeline’s final step significantly impacts our object-tracking system’s overall computation time.
While we have already filtered the object point cloud $\objectpointcloud$ to retain only the points associated with dynamic objects, the number of points can still be substantial, particularly in scenarios with many or large objects in the scene.

Modern lidar sensors produce dense point clouds, where neighboring points often convey redundant information.
To address this, we apply a widely adopted voxelization technique from the pcl library \cite{rusu2011pcl}.
Voxelization groups nearby points into a single representative point, effectively reducing the resolution of the point cloud while preserving its essential spatial structure.
We refer to the result of this process as the approximate object point cloud $\approxobjectpointcloud$.
The voxelization filter’s leaf size, which defines the resolution of the voxel grid, has significant impact on balancing the trade-off between computational efficiency and the fidelity of the object representation.
We found that a leaf size of \SI{0.1}{\meter} retains enough information to reliably track traffic participants. 

\subsection{Object Detection}
\label{sec:object_detection}

The object detection process begins by using the approximate object point cloud $\currentapproxobjectpointcloud$ from the current time step.
The algorithm then applies Euclidean clustering to create clusters $\cluster_{j} \subseteq \currentapproxobjectpointcloud$.
Euclidean clustering is a fast algorithm that groups nearby points into clusters based on spatial proximity.
We select a cluster threshold that ensures each cluster contains points from only a single object.
While this approach effectively eliminates clusters that contain points from multiple objects, it may generate multiple clusters corresponding to the same object $ \object_{i} $.

To resolve this issue, we introduce an additional merging step.
We assume that all clusters corresponding to the same object $ \object_i $ lie within its bounding box.
By merging all clusters inside a bounding box, we form a single, unified cluster.
The algorithm follows these steps: First, we take the list of objects $ \object_i \in \mathcal{O}_{k-1} $ from the previous time step $ t_{k-1} $ and predict their motion into the current time step using a motion model (see \cref{sec:state_fusion}).
 
Next, we identify all clusters $ \cluster_{j} $ that fall within the bounding box $ \boundingbox_i $ of any predicted $ \object_i $ and merge them into a single cluster.
Clusters outside of any bounding box remain unchanged.

After merging or leaving clusters unchanged, we transform all clusters into oriented bounding boxes.
Several algorithms can generate axis-aligned or oriented bounding boxes.
Axis-aligned bounding boxes are fast and easy to compute but lack orientation information required for tracking larger objects like cars or trucks.
However, they suffice for tracking objects where orientation is difficult to determine from the point cloud, such as pedestrians.

Our approach uses a multi-model method that combines L-Shape and Cylinder matching.
We create oriented bounding boxes for all clusters or cluster sets using either L-Shape matching or Cylinder matching.
For L-Shape matching, we utilize the Autoware implementation \cite{kato2018autoware}, which references the work of \cite{zhang2017efficient}.
Cylinder matching applies when the object is classified as a pedestrian or when its square-equivalent circle diameter is smaller than a predefined threshold $ d_{min} $.

L-Shape matching leverages the orientation $ \theta_j $ of the associated object $ \object_j $ to fit the shape.
The algorithm optimizes orientation within the range $ \theta_j \pm \Delta\theta_{detection} $.
For non-associated clusters, we apply the same detection step but use default initialization parameters.
Specifically, Cylinder matching assigns the object an orientation of $ \theta_i = 0^{\circ} $, while L-Shape matching assumes $ \theta_i \in [0^{\circ},90^{\circ}] $ with $ w_i \leq l_i $.

The final step fills the association matrix $ \associationmatrix_{a,k} $ with detections from the shape-matching process that have already been associated with an object $ \object_i $.
The object detection step concludes with the partial association matrix $ \associationmatrix_{a,k} $ and a list of detected bounding boxes $ \boundingboxlist_j $ from the current time step.

\subsection{Association}
\label{sec:association}

The association step is applied to all bounding boxes $\boundingbox_i$ of clusters $\cluster_{j} \subseteq \tilde{\mathcal{P}}_{\mathcal{O},k}$ that our intersecting bounding box approach in the previous step did not associate.
These cases include newly detected objects, false positives, or objects that moved by more than their bounding box size within a single time step, making them ineligible for association via bounding box intersection.

The association process consists of three main steps.
First, we compute a hypothesis in form of a cost matrix $\costmatrix_k$ using a suitable distance measure.
Next, we convert $\costmatrix_k$ into a boolean assignment matrix $\associationmatrix_{b,k}$ mapping detections to global objects.
Finally, we apply a gating step to $\associationmatrix_{b,k}$ using $\costmatrix_k$ and a threshold $\costmatrixthreshold$, ensuring we retain only feasible associations.

To construct the cost matrix $\costmatrix_k$, we use the Euclidean distance between objects $\object_i$ and bounding boxes $\boundingbox_j$ of clusters $\cluster_{j}$.
However, Euclidean distance alone often fails to achieve sufficient performance.
More robust alternatives provide better results, such as the Mahalanobis distance or a handcrafted piecewise distance measure.
In our implementation, only a small percentage ($\text{MISSING}\%$) of associations rely on Euclidean distance in this step, while the bounding box intersection algorithm determines the majority.
Consequently, Euclidean distance serves primarily as a fallback method when the primary association metric fails, and no complex distance measure is required.

We employ the Hungarian algorithm to complete $\associationmatrix_{b,k}$ based on $\costmatrix_k$ to finalize the associations.
This algorithm ensures optimal one-to-one matches between objects $\object_j$ and bounding boxes $\boundingbox_i$.
However, since the cost matrix $\costmatrix_k$ is fully populated, it may include unfeasible associations.
We apply a gating step that filters $\associationmatrix_{b,k}$ using a cost threshold $\costmatrixthreshold$, eliminating unfeasible matches.

The final association matrix $\associationmatrix_k$ is constructed as a block matrix using the previously computed association matrices $\associationmatrix_{a,k}$ and $\associationmatrix_{b,k}$:  

\begin{equation}
    \associationmatrix_{k} = \begin{bmatrix}\associationmatrix_{a,k} & \mathbf{0} \\ \mathbf{0} & \associationmatrix_{b,k}\end{bmatrix}
\end{equation}

\subsection{State fusion}
\label{sec:state_fusion}

Our state estimation uses an adapted Kalman filter.
Unlike classical implementations, we split the prediction and update steps, introducing additional steps in between.
The prediction step applies the motion model at the beginning of the tracking process to estimate the object's current state.
Later, the Kalman filter update step fuses detections with existing objects.

Lidar-based bounding box estimation provides limited information about the state space.
The available information is expressed in the measurement state space as:  
\begin{equation}
\measurement=\left[\positionvector,\yaw\right]^T
\end{equation}
where $ \positionvector=\left[\positionx, \positiony \right]^T $ and $ \yaw $ remain consistent with the state definition.
 
This leads to the following measurement mapping:  
\begin{equation}
\measurement = \measurementmatrix \state, \quad \measurementmatrix=\begin{bmatrix}1 & 0 & 0& 0& 0& 0 \\ 0 & 1 & 0& 0& 0& 0 \\ 0 & 0 & 0& 0& 1& 0\end{bmatrix}
\end{equation}

Since the measurement space lacks dynamic information, we model objects as extended and oriented point masses and use a constant-acceleration motion model.
 
More complex motion models typically cater to specific object types, such as cars, pedestrians, or cyclists.
However, robust real-time classification from lidar point clouds is challenging and would possible exceed the available computation power/time, so we opt for a simpler model that allows for class independent prediction.

Our motion model is defined as:  
\begin{equation}
\state_k = \statetransitionmatrix_k \state_{k-1} + \stateinputmatrix \stateinput_k + \omega_k, \quad \omega_k \sim \mathcal{N}(0,\processnoisematrix)
\end{equation}
where  
\begin{equation}
\state = \begin{bmatrix} \positionx \\ \positiony \\ \velocityx \\ \velocityy \\ \yaw \\ \yawrate \end{bmatrix}, \quad
\statetransitionmatrix = \begin{bmatrix}1 & 0 & dt & 0 & 0 & 0\\ 0 & 1 & 0 & dt & 0 & 0 \\ 0 & 0 & 1 & 0 & 0 & 0 \\ 0 & 0 & 0 & 1 & 0 & 0 \\ 0 & 0 & 0 & 0 & 1 & dt \\ 0 & 0 & 0 & 0 & 0 & 1\end{bmatrix}
\end{equation}

\begin{equation}
\stateinput = \begin{bmatrix} \accelerationx \\ \accelerationy \\ \yawraterate \end{bmatrix}, \quad
\stateinputmatrix = \begin{bmatrix} \frac{dt^2}{2} & 0 & 0\\ 0 & \frac{dt^2}{2} & 0 \\ dt & 0 & 0 \\ 0 & dt & 0 \\ 0 & 0 & \frac{dt^2}{2} \\ 0 & 0 & dt \end{bmatrix}
\end{equation}

We modeled the process noise covariance matrix using a discretized Wiener process model:

\begin{equation}
\label{eq:process_noise}
Q = \stateinputmatrix \pmb{W} \stateinputmatrix^T, \quad \pmb{W}=\begin{bmatrix} q_{a_x} & 0 & 0\\ 0 & q_{a_y} & 0 \\ 0 & 0 & q_{\dot{\yawrate}} \end{bmatrix}
\end{equation}

Our motion model requires estimating $ \stateinput $, which cannot be directly measured from $ \approxobjectpointcloud $.
We derive this estimate using tracking history by storing state estimations over the last $ t_{history} $ seconds and computing the acceleration input based on past observations.
If the tracking history lacks sufficient data, we set $ \stateinput = \mathbf{0} $, effectively defaulting to a constant velocity motion model.
This approach ensures stability when motion data is limited while allowing the model to adapt when enough history is available.

\subsection{Dimension Estimation}
\label{sec:dimension_estimation}

Classical dimension estimation using lidar relies on detecting the L-shape of an object to determine its dimensions.
This method performs well when the entire L-shape is visible but struggles near the edges of the FOV of the sensor.
Additionally, our aggressive preprocessing pipeline, which employs a fast clustering algorithm, can yield noisy results, and occasionally incorrect bounding box detections.
These issues lead to situations where the covariance of the dimension estimate appears low because the L-shape fits well, even when the clustering fails to separate multiple objects properly, making the estimator not perform as expected.

We aim to develop a fast estimation method that is robust against occasional wrong dimension estimation with high confidence values.
To achieve this, we evaluated three approaches: a handcrafted rule-based method based on weighted averages, a Kalman filter approach, and a grid-map-based method.
\begin{enumerate*}
    \item The rule-based method produced acceptable results but could not express the current confidence in the dimension estimation.
Possible downstream algorithms require a confidence value, such as a probabilistic confidence metric or variance, mainly when fusing the object list with other data sources.
    \item The Kalman filter approach also performed adequately but was highly sensitive to the clustering issues mentioned earlier.
Incorrect clustering often led to erroneous dimension estimates with low variance, falsely indicating high confidence.
The Kalman filter trusts this value greatly, and the dimension diverges from the actual value.
    \item The grid-map-based method we present below addresses both limitations.
It provides a variance estimate and remains robust against occasional erroneous measurements.
\end{enumerate*}

The grid-map-based approach, originally developed by Aeberhard in \cite{aeberhard2017object}, assumes independence between the object's length, width, and height measurements.
We discretize each dimension of the 3D bounding box into a one-dimensional grid map.
Each grid divides the dimension into cells of equal size, with a step size $\Delta d^i$ ranging from $\SI{0.0}{\meter}$ to the maximum size of an object in the given dimension, denoted as $d_{max}^i$.
The dimension $i$ refers to length ($l$), width ($w$), or height ($h$).
The midpoint of each cell $j$ in the grid is represented by $d_j^i$.

Choosing the step size $\Delta d^i$ involves a trade-off between accuracy and computational effort.
From our tests, we found that setting $\Delta d^i = \SI{0.1}{\meter}$ produced satisfactory results.
Each cell also stores the log-odds ratio, $\mathcal{L}_j^i$, representing the cumulative probability that the object is at least as large as $d_j^i$.

We convert the log-odds ratio into cumulative probability using the equation:
\begin{equation}
    p(d^i \leq d^i_j) = \frac{1}{1 + \exp\left(\mathcal{L}_j^i\right)}.
\end{equation}

To calculate the likelihood $p(d^i_j)$ of an object having a dimension of $d^i_j$, we compute finite differences from the cumulative probability.

Finally, we estimate the object's dimension $\tilde{d}^i$ and its standard deviation $\sigma_{\tilde{d}^i}$ using the weighted sum:
\begin{align}
    \tilde{d}^i =& \sum_j p(d^i_j) \cdot d^i_j \\
    \sigma_{\tilde{d}^i} =& \sqrt{\sum_j p(d^i_j) \cdot \left(d^i_j - \tilde{d}^i\right)^2} 
\end{align}
    
The algorithm involves several key steps to ensure robust dimension estimation.
We start by initializing the grid cells with an initial value.
For a binary Bayes filter, this is typically done using  
\begin{equation}
    \mathcal{L}_j^i = \log \left( \frac{P_{ref}}{1-P_{ref}} \right), \forall i,j.
\end{equation}
Setting $P_{ref} = 0.5$ simplifies this step, as it zero-initializes all cells, a reasonable and effective assumption.

To update the grid cells, we use the measurements $\measureddimension = \left(\hat{l}, \hat{w}, \hat{h}\right)^T$ and the associated covariance matrix:  
\begin{equation}
    \measureddimensioncovariance =  
    \begin{bmatrix}
        \sigma_{\hat{l}} & 0 & 0\\
        0 & \sigma_{\hat{w}} & 0 \\
        0 & 0 & \sigma_{\hat{h}}
    \end{bmatrix}.
\end{equation}

For individual dimensions, we rely on scalar values $\measureddimensionelementi$ and $\measureddimensionelementivariance$.
We then apply a binary Bayes filter to update the log-odds, $\mathcal{L}_j^i$, using the likelihood $p_d\left(d^i_j | \{\measureddimensionelementi, \measureddimensionelementivariance\}\right)$.

The original function for this likelihood, proposed by Aeberhard, is not continuous and contains gaps.
While adequate for its original use, we require a smoother, universal approach.
To address this, we define:  
\begin{equation}
p_d\left(d^i_j | \{\measureddimensionelementi, \measureddimensionelementivariance\}\right) = \left\{
\begin{array}{ll}
      \frac{-2 \left( P_{max} - P_{ref}\right)}{1+\exp\left( \frac{1}{\sigma_{\tilde{d}^i}} \cdot \left(\tilde{d}^i - d_j^i\right)\right)} + P_{max}, & \tilde{d}^i < d^i_j, \\
      P_{ref}, & \tilde{d}^i = d^i_j, \\
      \frac{- 2 P_{ref}}{1+\exp\left( \frac{1}{\sigma_{\tilde{d}^i}} \cdot \left(\tilde{d}^i - d_j^i\right)\right)} + 2 P_{ref}, & \tilde{d}^i > d^i_j.
\end{array} 
\right.
\label{eq:p_d}
\end{equation}
This function ensures smooth behavior for all input sizes and conditions, addressing the limitations of its predecessor.

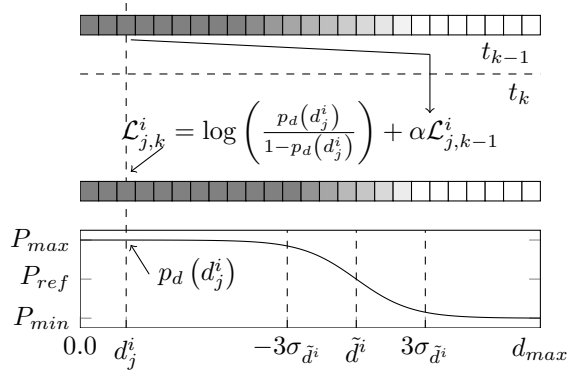
\begin{figure}[tbp]
    \centering
    \input{dimension_estimation}
    \caption{The illustration depicts the 1D gridmap approach to updating the dimension estimation.
The function $p_d$ is parameterized based on the current measurement and variance of the dimension.}
    \label{fig:discretization_dimension}
\end{figure}

Finally, as shown by \cref{fig:discretization_dimension}, we update the log-odds for each cell using the Bayes filter rule:  
\begin{equation}
    \mathcal{L}^i_{j,k} = \log \left( \frac{p_d\left(d^i_j\right)}{1-p_d\left(d^i_j\right)} \right) +  \alpha \mathcal{L}^i_{j,k-1}.
\end{equation} 
Here, $ \alpha $ acts as a forgetting factor, an addition not present in the original work, allowing the estimated dimensions to adapt to consistent measurements indicating a change in size, whether growth or shrinkage, while resisting the influence of isolated erroneous measurements.
Omitting $ \alpha $ or setting it too high causes the algorithm to lock onto a dimension estimate after observing a sufficient amount of information.
While desirable for certain applications, this prevents objects from expanding at the edges of the sensor's FOV.
We observe that setting $\alpha = 0.92$ enables objects to shrink and grow at an adequate rate and setting $\alpha >= 0.95$ will start to show the locking behavior.
 
An example of the described behaviors is illustrated at the top of \Cref{fig:length_classification} showing that the algorithm remains robust under against occasional wrong bounding box estimation.
The results from the illustration can be reproduces using our Matlab example (see \nameref{sec:openmaterial})

\subsection{Existence probability estimation}
\label{sec:existence_probability_estimation}

This subsection presents a heuristic-based approach for computing an object’s existence probability \(\existenceprobability\).
The algorithm aggregates six weighted factors, as follows:

\begin{equation}
    \existenceprobability = \sum_j w_j \cdot f_j,
\end{equation}

where \(w_j\) are the weights, and \(f_j\) are the existence factors.
Specifically, the current implementation employs:
\[
f_{N}, f_{A}, f_{AR}, f_{V}, f_{\velocityvector}, f_{\yawrate}.
\]

The first two factors measure the track’s history.
The term \(f_{N}\) uses a logistic function to convert the track’s age \(N\) into a score:

\[
    f_{N} = \frac{1}{1 + \exp\bigl(-k_{\mathrm{age}}(N - N_{\mathrm{offset}})\bigr)},
\]

where \(k_{\mathrm{age}}=0.4\) and \(N_{\mathrm{offset}}=5\) have shown favorable performance.
The second factor, \(f_{A}\), captures the fraction of frames in the track history with successful associations, thus penalizing objects that often lack measurement updates.

The next two factors focus on bounding box characteristics.
Factor \(f_{AR}\) incorporates the standard deviation \(\sigma_{AR}\) of the bounding box’s aspect ratio:

\begin{equation}
    f_{AR} = 1 \;-\; \frac{\sigma_{AR}}{t_{AR}},
    \label{eq:existance}
\end{equation}

where \(t_{AR}=1.0\).
A larger \(\sigma_{AR}\) (i.e., rapid changes in the bounding box’s shape) reduces the factor value, thereby lowering the overall existence probability.
A similar formulation applies to \(f_{V}\), which uses the standard deviation of the bounding box’s volume in place of \(\sigma_{AR}\), with \(t_{V} = 1.0\).

The last two factors, \(f_{\velocityvector}\) and \(f_{\yawrate}\), quantify changes in the state vector.
Each applies the same functional form as \eqref{eq:existance} while monitoring the variability in velocity and yaw rate, respectively.
Thresholds of \(t_{\velocityvector}=2.0\) and \(t_{\yawrate}=1.0\) penalize large fluctuations, reflecting the notion that abruptly changing motion profiles cast doubts on a track’s authenticity.

Although no parameter optimization has been performed in the current implementation, preliminary results in \cref{sec:results} demonstrate that this heuristic yields favorable performance for existence estimation.

\subsection{Classification}
\label{sec:classification}

Traditionally, classification methods analyze the shape of a point cloud using machine learning techniques to determine the class of an object in the point cloud.
However, these methods often have significant limitations:
\begin{enumerate*}
    \item They are typically not real-time capable.
    \item They require dense and well-defined point clouds, which are not available the our approach.
\end{enumerate*}
To address these issues, we propose a fast and straightforward probability-based classification approach.
Our method utilizes a probaility-based lookup table to provide a real-time estimation of the object type.

We build upon the probability-based dimension estimation method described in \cref{sec:dimension_estimation}.
Our approach computes a probability value for each object class based on independent dimension estimations.
Specifically, the estimation for each dimension $i$ (length $l$, width $w$, or height $h$) is treated independently.

Given the estimated value $\tilde{d}^i$ and its associated standard deviation $\sigma_{\tilde{d}^i}$ for each dimension, the goal is to compute the probability that the object belongs to a specific class $\classification_j$.
The results populate the classification vector $\classificationvector$, which contains the probability values for all considered classes.

For each class $\classification_j$, the algorithm requires the mean dimension $\mu^i_j$ and standard deviation $\sigma^i_j$, which model the normal distribution of dimension $i$ within class $j$.
These values can be determined using official records of registered vehicles in the area of interest.
For example, the open Canadian Vehicles in Circulation database provides a comprehensive list of all registered vehicles\footnote{https://open.canada.ca/}.
However, this approach only accounts for vehicles requiring registration and does not include pedestrians, bicycles, e-scooters, or other unregistered objects.

Alternatively, the algorithm can fit these distributions to an existing labeled dataset.
We used the Level-X datasets\cite{highDdataset, inDdataset, exiDdataset, unidDataset, rounDdataset}, but any dataset with sufficient labeling can serve the same purpose.
The Level-X datasets provide length and width estimations of objects but lack height information.
To address this limitation, we assume $ p(\classification_j | \estimateddimensionelement^h ) = 1\ \forall\ j $.
Using these labeled datasets results in the classification vector $\classificationvector=\{c_{pedestrian}, c_{cyclist}, c_{motorcycle}, c_{car}, c_{truck}, c_{other}\}$, where the "other" category captures any remaining unclassified objects.

The classification probability for class $\classification_j$ is computed as:
\begin{equation}
    p(\classification_j | \estimateddimension ) = \prod_i p(\classification_j | \estimateddimensionelement^i ).
\end{equation}

Using Bayes' theorem, the probability $p(\classification_j | \estimateddimensionelement^i )$ can be expressed as:
\begin{equation}
    p(\classification_j | \estimateddimensionelement^i ) = \frac{p(\estimateddimensionelement^i | \classification_j)p(\classification_j)}{p(\estimateddimensionelement^i)}.
\end{equation}

The two missing terms in the equation are defined as:
\begin{itemize}
    \item The likelihood of the measurement given the class:
    \begin{equation}
        p(\estimateddimensionelement^i | \classification_j) = \frac{1}{\sqrt{2\pi \sigma_\text{combined}^2}} \exp\left(-\frac{(\estimateddimensionelement^i - \mu^i_j)^2}{2\sigma_\text{combined}^2}\right),
    \end{equation}
    where $\sigma_\text{combined}^2 = {\sigma^i_j}^2 + \sigma_{\estimateddimensionelement^i}^2$.

    \item The marginal likelihood:
    \begin{equation}
        p(\estimateddimensionelement^i) = \sum_{j} p(\estimateddimensionelement^i | \classification_j) p(\classification_j).
    \end{equation}
\end{itemize}

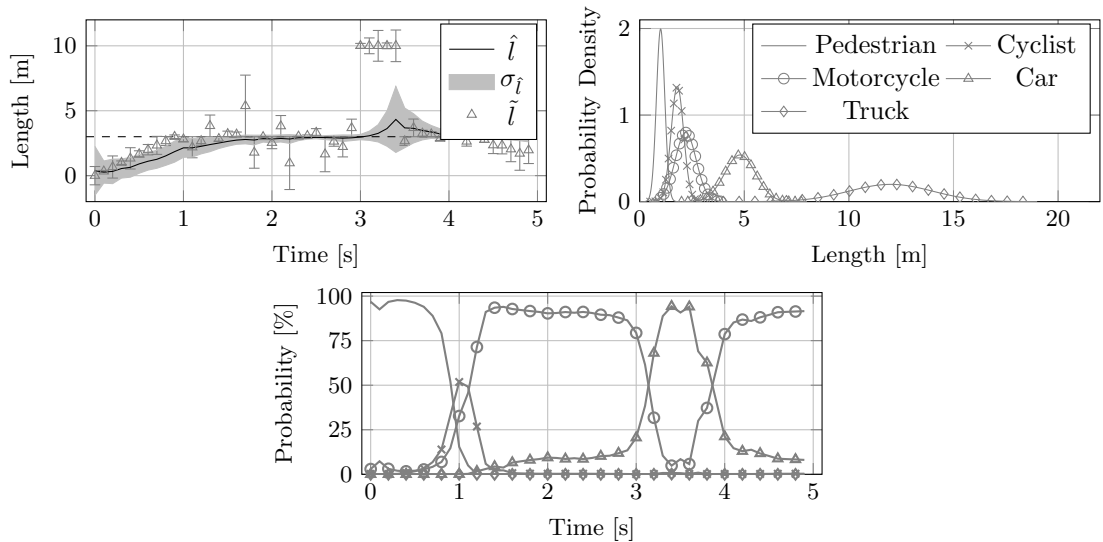
\begin{figure}[tbp]
    \centering
    \input{length_estimation}
    \input{length_classification}
    \input{classification_estimation}
    \caption{Example of dimension-based classification and estimation in three stacked plots.
The top plot demonstrates the length estimation process, revealing how the probabilistic approach converges toward the true value while remaining robust to large deviations.
The middle plot illustrates length-based classification with five candidate classes.
The bottom plot shows how classification probabilities evolve over time, using estimates from the top plot and demonstrating temporal fluctuations as the algorithm refines its dimension estimate.}
    
    \label{fig:length_classification}
\end{figure}

Figure \ref{fig:length_classification} illustrates the classification results obtained using length estimation. The length estimation from Section \ref{sec:dimension_estimation} is compared against the class probabilities at each timestep, leading to a continuous class estimation. Notably, this classifier is a single-shot classifier with no memory. The temporal memory results from the behavior of the dimension estimator, which is recursive and relies on the previous timestep.

\subsection{Object management}
\label{sec:object_management}

The object management step marks the final stage of our proposed algorithm.
It combines the deletion of outdated objects and the initiation of new ones.
We initiate new objects whenever a measurement cannot be associated with any existing object in the current object list.
To initiate an object, we define its parameters using the detected bounding box as $O_{init} = \boundingbox_i$.
State variables not defined by the bounding box are set as follows: the velocity vector is $\velocityvector = \mathbf{0}$, the yaw rate is $\yawrate = 0$, the existence probability is set to $\existenceprobability = 0.5$, indicating an undecided probability, and the classification vector is $\classificationvector = \{0, 0, 0, 0, 0, 1\}$, indicating an unclassified object.

The algorithm deletes objects that remain unobserved for $t_{delete}(\positionvector)$ seconds. This threshold is determined by the position of the sensor relative to the object. We use a static map that indicates statically occluded and statically visible areas. Occluded areas allow for a higher deletion threshold because the object might not be continuously trackable in those areas. Additionally, these areas prevent the creation of new objects. Visible areas have the normal deletion threshold, and areas that are neither on the map nor in the point cloud are deleted immediately by setting the threshold to $t_{delete} = \SI{0}{\second}$. The presented algorithm uses a threshold of $t_{delete} = \SI{1}{\second}$, which can be extended up to $t_{delete} = \SI{2}{\second}$ in occluded areas.
Currently, the maps are manually created for each location. However, given the technical specifications of the sensor and the ground points removed in \cref{sec:point_cloud_preprocessing}, it should be possible to automatically initialize and update those maps using the point cloud. This approach is left for further research.

 
\section{Experimental results and discussion}
\label{sec:results}

\begin{figure}[tbp]
    \centering
    \includegraphics[width=8cm]{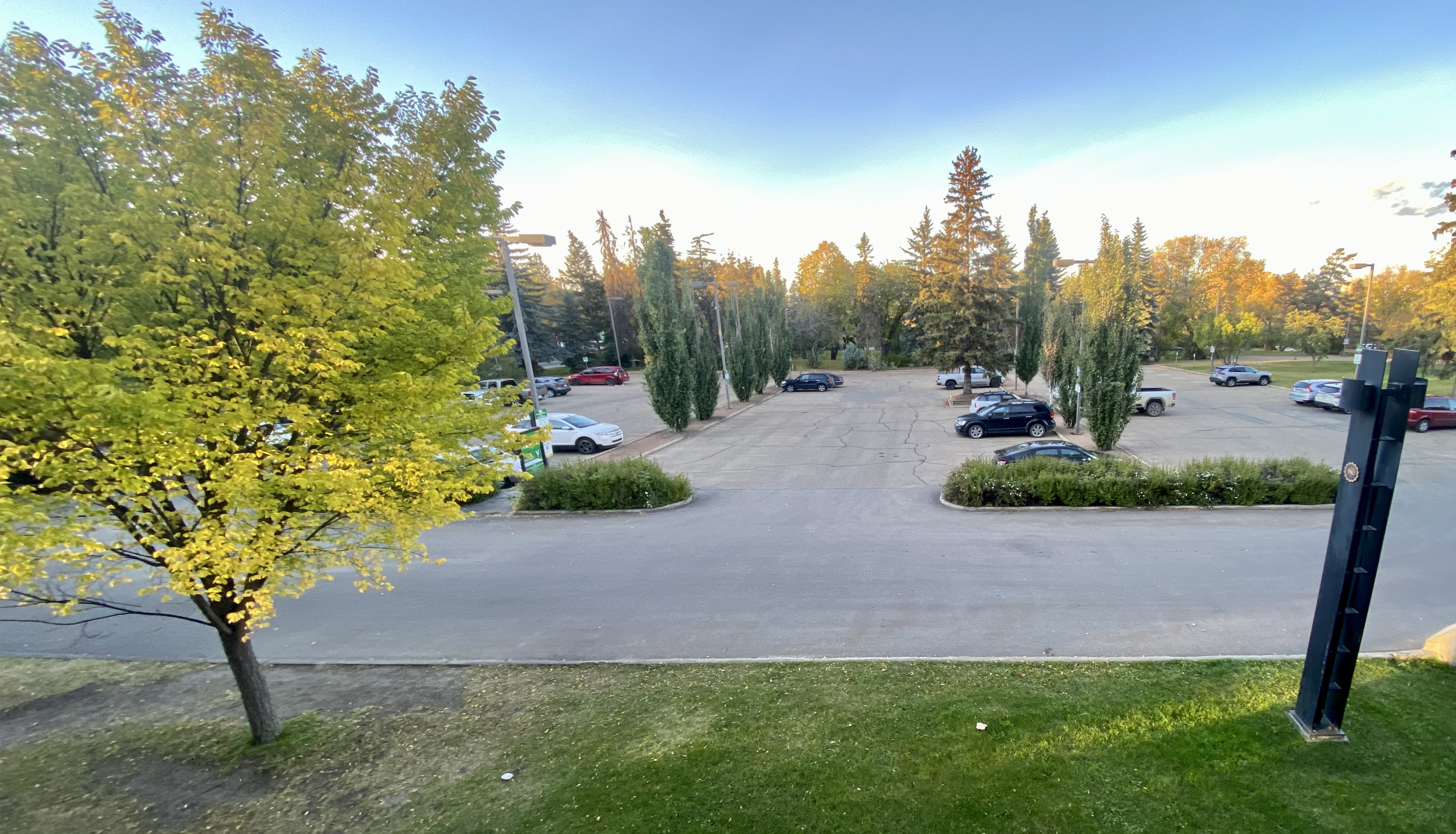}
    \includegraphics[width=8cm]{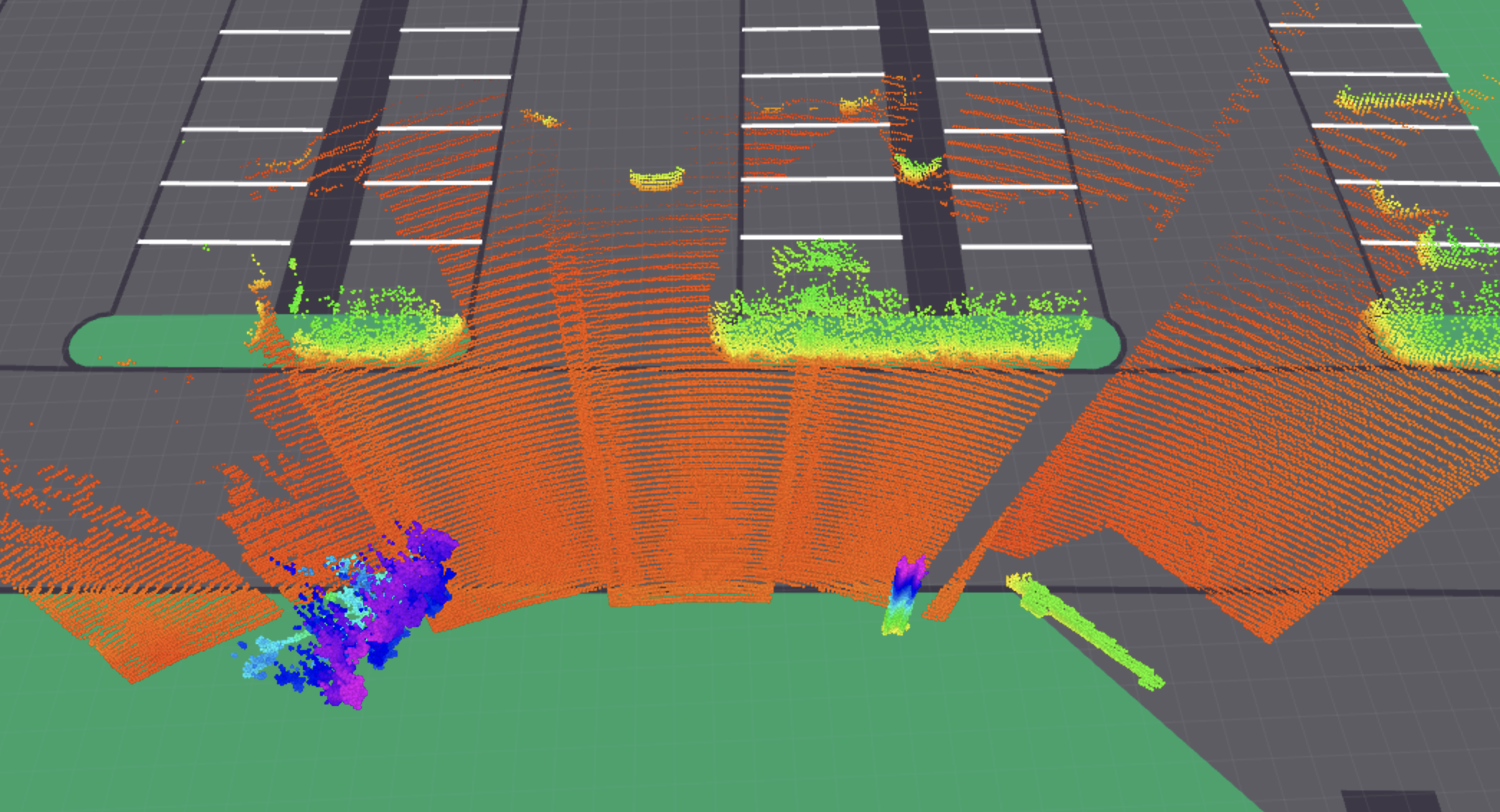}

    \caption{This picture depicts the testing site on the University of Alberta campus including a raw point cloud.
The site features both semi-static objects, such as the tree on the left and the vegetation in the background, and static objects, like the light pole on the right.
The suboptimal environment for lidar tracking arises from the necessity of utilizing existing infrastructure to mount the sensor system.}
    \label{fig:testing_site}
\end{figure}

Experiments take place on the University of Alberta campus, utilizing a site characterized by semi-static vegetation and static objects such as light poles.
This environment reflects realistic mounting conditions, where existing infrastructure often dictates sensor placement and leads to suboptimal performance.
\Cref{fig:testing_site} illustrates the site and a raw lidar point cloud, highlighting heavy occlusions caused by vegetation on the left and static objects on the right.
These occlusions create persistent blind spots throughout all recorded frames.

The lidar unit is mounted at a height of \SI{5.9}{\meter}, pitched downward by \SI{20}{\deg}, and rolled by \SI{180}{\deg} to access mounting holes.
Due to the sensor’s symmetrical vertical field of view, this roll angle does not reduce coverage.
The sensor remains aligned with zero yaw relative to the local map coordinate system, offers a maximum range of \SI{30}{\meter}, and provides a \SI{120}{\deg} horizontal field of view at an update rate of \SI{10}{\hertz}.

Data collection comprises two scenarios, each lasting approximately five minutes and capturing around 3{,}100 lidar frames under favorable summer weather and lighting conditions.
A test vehicle equipped with RTK GPS (accuracy of around \SI{5}{\centi\meter}) supplies ground-truth measurements for dynamic objects, and a nearby parking lot enables evaluation of the tracking algorithm on static objects.

Our evaluation address three primary metrics: object existence, tracking accuracy (state and dimension), and classification accuracy.
The object existence metric considers how frequently the system misses an object, falsely generates an object (false postitives), or experiences ID switches during tracking.
The tracking accuracy metric measures how closely the estimated state and dimensions of the lidar align with ground-truth data from the test vehicle.
Finally, the classification accuracy metric verifies how reliably the system assigns the correct class to each object in the frame, based on manual annotations.

\subsection{Object Existence}
\label{sec:results_object_existence}

\begin{figure}
    \centering
    \input{existence_histogram}
    \caption{
            Distribution of mean existence probabilities of all tracks.
The plot indicated a clear separation between true positives and false positives.
        }
    \label{fig:existence_estimation}
\end{figure}
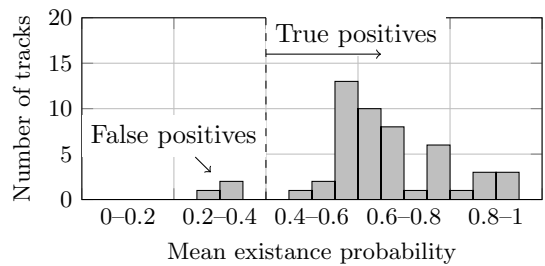

We collected a total of 51 distinct tracks during the experiment.
The presented algorithm produced three false positives, two classified as cars and one as a motorcycle.
All three false positives where triggered by vegetation during the simulated wind conditions.
We shook the sensor with different amplitudes, until the static object removal malfunctioned.
We observer acceleration of up to $\SI{2.0}{\meter\per\second^2}$ at sensor level.
Removing these false positives would require a compensation algorithm that corrects for the movement of the infrastructure on which the sensor is mounted.
These false positives did not occur during normal operation.
Each incorrect detection yielded an mean existence probability below 35\%, whereas genuine objects consistently maintained mean probabilities above 50\%.
Figure~\ref{fig:existence_estimation} illustrates this clear gap, with the three false positives clustering at the lower end of the existence probability histogram.
We did not miss any dynamic objects across all recordings, both during wind simulation and under normal conditions.
However, one track broke and split into two separate objects, and only one continued to be tracked.
This issue resulted from the current limitation of the association algorithm, which assumes a strict one-to-one mapping between detections and global objects.

\subsection{Track Accuracy}
\label{sec:results_track_accuracy}

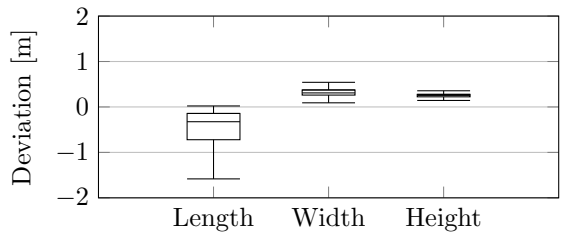
\begin{figure}[tbp]
    \centering
    \input{error_dimension}
    \caption{
        Tracking deviation of the detected vehicles showing the deviation in dimension estimation.
		}
    \label{fig:dimension_estimation}
\end{figure}

This section evaluates two primary components of tracking accuracy: the pose (position and orientation) and the estimated dimensions of the objects.
Both assessments rely on the RMSE as a global indicator, complemented by box plots for statistical analysis of the data distribution.
The dataset comprises 19 unique passes of the measurement vehicle, where the car travels from right to left, from left to right, and toward the sensor node.
Other maneuvers were not feasible due to the road configuration at the test site.
These 19 passes generated 1,736 unique data points for comparison.
Figure~\ref{fig:dimension_estimation} depicts box plots for both pose and dimension estimations.

The pose estimation exhibits an RMSE of \SI{0.52}{\meter}, which depends strongly on dimension estimates with an RMSE of \SI{0.46}{\meter}.
The algorithm defines the pose as the center of the object’s bounding box, introducing a correlation between bounding box errors and position errors.
In particular, the observed underestimation of the object’s length and overestimation of its width significantly contribute to the overall pose error.
Moreover, the tracker’s performance degrades noticeably when the L-shape of the object cannot be clearly identified in the point cloud, a behavior also noted in related research.

Orientation estimation proves challenging, given a RMSE of \SI{7.69}{\deg},  especially near the edges of the lidars’s field of view.
When an object partially appears in the sensor’s peripheral range, the bounding box orientation remains ambiguous until the object fully enters the coverage.
Incorporating HD map information could help infer the most probable orientation of vehicles as they enter or exit the scene, given that road users typically align themselves with existing lanes.

Dimension estimation tends to underestimate the size of objects.
This bias may be mitigated by refinements to \eqref{eq:p_d}, particularly through a more asymmetrical probability distribution.
The current approach treats under- and overestimation with equal weight, but a one-sided function—such as a Heaviside-like step—may better reflect the consistent underestimation observed in practice.

\begin{figure}[tbp]
    \centering
    \input{error_tracking}
    \caption{
        Tracking deviation of the detected vehicles showing the position deviation along the sensors forward/lateral direction, and orientation deviation.
		}
    \label{fig:position_estimation}
\end{figure}
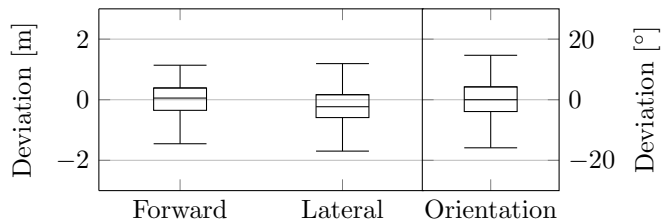

\subsection{Object Classification}
\label{sec:results_object_classification}

A total of 51 distinct tracks, minus the three false positives, yielded 48 valid tracks.
We used manual labeling to identify the correct class for every object in the dataset.
After establishing ground-truth labels for all object IDs, we compared those labels to the classifications assigned by the algorithm for the majority of each track’s duration.
In \cref{fig:length_classification}, for instance, the classifier would assign the “motorcycle” category if that label dominated most of the track’s frames.

Among the 48 valid tracks, 47 were correctly classified.
Only one object, a cyclist, was misclassified as a pedestrian when moving directly toward the lidar, resulting in a smaller footprint that statistically resembled a pedestrian.
This issue arises because a cyclist and a pedestrian often share similar widths.
Revisiting \cref{fig:length_classification}, the classifier transitions among multiple classes while the dimension estimate increases.
Introducing additional temporal memory could address this classification drift by prioritizing the most probable class over the object’s track history.

In all 19 passes of the reference car, the classifier correctly identified its class, with 99.6\% of frames labeled as “car.” The remaining fraction of frames corresponded to moments when the car entered the field of view and the dimension estimate was still adjusting from its initial bounding box.

\subsection{Runtime}
\label{sec:results_runtime}

We use an NVIDIA Jetson Orin NX with 16GB of RAM and an 8-core Cortex-A78AE v8.2 64-bit CPU as the edge processing unit.
To evaluate runtime, we measure the time difference between the lidar driver’s timestamp and the arrival of the final object list message, a duration that encompasses both processing and communication overheads across all nodes.
Since the liadr sensor operates at \SI{10}{\hertz}, the system must complete data reception, processing, and message publication within \SI{100}{\milli\second}.

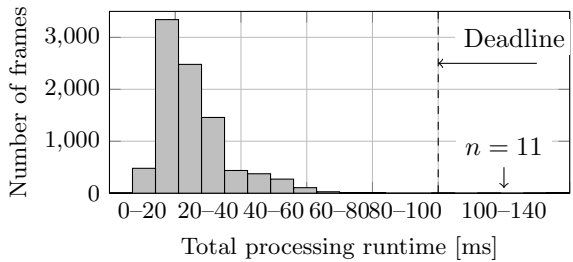
\begin{figure}
    \centering
    \input{timing_histogram}
    \caption{
            Distribution of total runtime of preprocessing and tracking.
The sensor is running at \SI{10}{\hertz} resulting in a deadline at \SI{100}{\milli\second}.
        }
    \label{fig:runtime}
\end{figure}

Experimental results reveal that 99.88\% of the messages in our dataset meet this \SI{100}{\milli\second} deadline.
Only 0.12\% exceed the deadline.
Furthermore, 95.76\% of the messages complete processing within \SI{50}{\milli\second}, indicating a left-skewed distribution of overall latency (see \cref{fig:runtime}).
CPU load appears evenly distributed across all eight cores, peaking at about 40\% utilization, while memory usage remains around 3GB.
Although the tracker and the point cloud preprocessor run sequentially, multiple processes, together with ROS2’s publish/subscribe model, permit concurrent message handling.
Notably, the current implementation relies solely on CPU resources, consuming less than 50\% of the available computational capacity.
This configuration leaves sufficient headroom for additional or more complex applications in the future.

Compared to the related work presented in \cref{sec:related_work}, our approach achieves an average total runtime of \SI{25.5}{\milli\second}.
However, directly comparing this to literature is not possible, as most authors refrain from providing an analysis of total runtime and instead provide inference or detection timings, which do not include preprocessing, trakcing, and management.
Given the available runtime for those subsystems, we can assume that our approach outperforms most approaches regarding total runtime. 

While most related work does not specify the wattage used during their experiments, it can be safely assumed that the embedded platform we used, which draws a maximum of \SI{30}{\watt}, consumed significantly less power compared to any of the full-size GPUs used. 

\section{Conclusion}
\label{sec:conclusion}

This work presented a lidar-only roadside sensor node with optional off-grid capability and a CPU-only edge tracking pipeline.
The proposed method provides object poses, velocities, bounding boxes, classifications, and existence probabilities in real time.
Experimental results confirmed that a rule-based approach can reliably detect, track, and classify dynamic objects in lidar data, with false positives occurring only under simulated wind-induced shaking of the sensor node.
A runtime analysis demonstrated that an off-the-shelf NVIDIA Jetson device meets the computational requirements, highlighting the approach’s cost-effectiveness and resource efficiency for edge-based tracking of road participants.
While we acknowledge the limitations of our current dataset in terms of size and weather variability, this foundational work enables future research in low-power embedded hardware as edge devices.

We intend to expand testing across a broader range of scenarios and diverse weather conditions within Canadian cities.
Additionally, exploring the fusion of multiple sensor nodes and incorporating additional sensors will provide valuable insights into the scalability of our proposed system.
Our results indicate that the embedded hardware used has sufficient remaining computational resources for further processing capabilities.

This ongoing research aims to refine our understanding of tracking methodologies in real-world urban environments while focusing on appropriately sized hardware that allows for potential deployment in future applications.




\medskip
\textbf{Acknowledgements} \par 
We acknowledge the financial support for this project by the Collaborative Research Center / Transregio 339 of the German Research Foundation (DFG).

We acknowledge the financial support for this project by the Natural Science and Engineering Research Council of Canada, RGPIN-2020-05097.

\medskip

%
\bibliographystyle{MSP}
\bibliography{WileyMSPlibrary}








\begin{figure}[hbt]
\textbf{Table of Contents}\\
\medskip
  \includegraphics{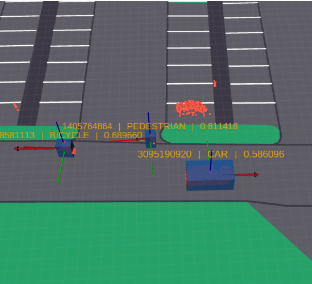}
  \medskip
  \caption*{This paper presents a lidar-based sensor node design and a rule-based state observer for edge-based traffic participant tracking. Unlike other state-of-the-art methods, this state observer enables real-time, CPU-only edge processing without relying on machine learning approaches. The evaluation indicates that the observer can accurately estimate the state, dimension, classification, and existence probability of traffic participants.}
\end{figure}

\end{document}

%% file: construction.tex
\begin{tikzpicture}[font=\small]
    \tikzstyle{field}=[text width=2.0cm, minimum height=1cm, minimum width=1cm, align=center, draw, inner sep=0.1cm, outer sep=0.1cm]
    \tikzstyle{blank_field}=[field, draw = none, minimum height=0.8cm]
    \tikzstyle{docker}=[blank_field, text width=3.0cm, align=left,minimum height=0.5cm,]

    \tikzstyle{field_arrow}=[->]
    \tikzstyle{model_arrow}=[->, dashed]
    
    \tikzstyle{power}=[dashed]
    \tikzstyle{data}=[]

    \tikzset{wireless/.pic={
        \draw [] (0,0) -| (.5,0.4);
        \node at (0.4,0.5) {\pgfuseplotmark{*}};
        \foreach \r in {.1,.2,.3}
            \draw (0.55,0.5) ++ (60:\r) arc (60:-60:\r);
    }}
    \tikzset{package/.pic={
        \draw [] (-0.15,0.1) -- ++(0.4,0.0) -- ++(-0.05,0.1) -- ++(-0.3,0.0) -- ++(-0.05,-0.1);
        \draw [] (-0.15,0.1) -- ++(0.0,-0.3) -- ++(0.4,0.0) -- ++(0.0,0.3);
        \draw [fill=black] (0.07,-0.12) rectangle ++(0.1,0.03);
    }}

    \tikzstyle{ground}=[fill,pattern=north east lines,draw=none]

    \def \vergap{0.5cm};
    \def \horgap{1.0cm};

    \draw[lightgray] (-1.9cm,-12.5cm) -- ++(0.0cm,12.5cm) -- ++(1.0cm,0.0cm) -- ++(0.0cm,0.25cm) -- ++(2.0cm,0.0cm) -- ++(0.0cm,-1.0cm) -- ++(-2.0cm,0.0cm) -- ++(0.0cm,0.25cm) -- ++(-0.55cm,0.0cm) -- ++(0.0cm,-12.0cm) -- ++(-0.45cm,0.0cm);

    \draw[pattern color=lightgray, ground] (-4.5cm,-12.5cm) rectangle ++(8.8cm,-0.3cm);
    \draw[lightgray] (-4.5cm,-12.5cm) -- ++(8.8cm,0.0cm);

    \node[blank_field] at (0,0) (AR) {};
    \node[field, below=2.0*\vergap of AR] (BR) {Lidar Sensor};
    \node[field, left=\horgap of BR] (BL) {Solar Panel};
    \node[blank_field, below=0.5*\vergap of BR] (CR) {};

    \node[field, below=2.0*\vergap of CR] (ER) {Sensor Interface};
    \node[blank_field, right=-0.2cm of ER] (ERR) {};

    \node[field, below=0.5*\vergap of ER] (FR) {Switch};
    \node[field, right=-0.2cm of FR] (FRR) {Router};

    \pic [right=-0.0cm of FRR]{wireless};
    
    \node[blank_field, below=0.5*\vergap of FR] (GR) {};
    \node[field, below left=0.5*\vergap and 0cm of GR, minimum width=4.4cm, anchor=north west, align=left, text width=4.2cm, text depth=2.6cm] (HR) {Edge Computation Unit};

    \node[docker,below right=0.8cm and 0.8cm of HR.north west] (HR1) {Sensor Driver};
    \pic [left=0.2cm of HR1]{package};

    \node[docker,below=0.0cm of HR1] (HR2) {Diagnostic Interface};
    \pic [left=0.2cm of HR2]{package};
    
    \node[docker,below=0.0cm of HR2] (HR3) {State Observer};
    \pic [left=0.2cm of HR3]{package};

    \node[field, left=\horgap of ER] (EL) {Charge Controller};
    \node[blank_field, above=2.0*\vergap of EL] (DL) {};
    \node[field, below=0.5*\vergap of EL] (FL) {Battery};
    \node[blank_field, left=\horgap of GR] (GL) {};
    \node[field, below=0.5*\vergap of GL] (HL) {12V Power Supply};

    \draw[data] ([xshift=0.1cm]ER.south) -- ([xshift=0.1cm]FR.north);

    \draw[power] (BL) -- (EL);
    \draw[data] ([xshift=0.1cm]BR.south) -- ([xshift=0.1cm]ER.north);
    \draw[power] ([xshift=-0.1cm]BR.south) -- ([xshift=-0.1cm]ER.north);

    \draw[power] (FL) -- (EL);
    \draw[power] (EL) -- (ER);
    \draw[power] ([xshift=0.5*\horgap]EL.east) |- (FR);
    
    \draw[power] (HL) |- ([xshift=-0.5*\horgap]GR.west);
    \draw[power] ([xshift=-0.5*\horgap]GR.west) -| ([xshift=-1.2cm]HR.north);

    \draw[power] ([xshift=-0.5*\horgap]GR.west) -- ([xshift=-0.5*\horgap]FR.west) ;

    \draw[data] ([xshift=0.1cm]FR.south) -- ([xshift=-1.0cm]HR.north);

    \node[below right=\vergap and -0.4cm of HL, anchor=north east] (Legend1) {Power};
    \draw[power] (Legend1.west) -- ([xshift=-0.5cm]Legend1.west) ;

    \node[below left=0.0cm and 0.0cm of Legend1, anchor=north west] (Legend2){Data};
    \draw[data] (Legend2.west) -- ([xshift=-0.5cm]Legend2.west) ;

\end{tikzpicture}

%% file: overview.tex
\begin{tikzpicture}[font=\small]
    \tikzstyle{field}=[text width=3cm, minimum height=1cm, minimum width=1cm, align=center, draw, inner sep=0.1cm, outer sep=0.1cm]
    \tikzstyle{sum}=[field, text width=0cm, rounded corners=0.25cm, minimum height=0.5cm, minimum width=0.5cm]
    \tikzstyle{model}=[field, tape, tape bend top=none, tape bend height=2mm, tape bend bottom=out and in]

    \tikzstyle{field_arrow}=[->]
    \tikzstyle{model_arrow}=[->, dashed]

    \def \vergap{0.5cm};
    \def \horgap{1.0cm};

    \node[] at (0,0) (AR) {};
    \node[field, below=2.0*\vergap of AR] (BR) {Ground removal};
    \node[field, below=0.5*\vergap of BR] (CR) {Static object removal};
    \node[field, below=0.5*\vergap of CR] (DR) {Voxelization};

    \node[right=0cm of CR, rotate=-90, anchor=south] {Point cloud preprocessing};
    \draw[dashed] (BR.north west) rectangle (DR.south east);

    \node[field, below=2.0*\vergap of DR] (ER) {Detector};
    \node[field, below=0.5*\vergap of ER] (FR) {Hypothesizer};
    \node[field, below=0.5*\vergap of FR] (GR) {Associator};
    \node[field, below=0.5*\vergap of GR] (HR) {Initiator};

    \node[field, left=\horgap of ER] (EL) {Predictor};
    \node[above=2.0*\vergap of EL] (DL) {};
    \node[model, below=0.5*\vergap of EL] (FL) {Motion model};
    \node[field, left=\horgap of GR] (GL) {Updater};
    \node[field, below=0.5*\vergap of GL] (HL) {Deleter};
    \node[field, fill=none, draw=none, below=0.5*\vergap of HL] (IL2) {};
    \node[below=2.0*\vergap of IL2] (JL) {};

    \node[sum] at (IL2.center) (IL) {};
    
    \node[right=0cm of GR, rotate=-90, anchor=south] {Object detector and tracker};
    \draw[dashed] (ER.north east) rectangle (IL2.south west);

    \draw[field_arrow] (AR) -- node [right] {$\currentpointcloud$} (BR); 
    \draw[field_arrow] (BR) -- (CR);
    \draw[field_arrow] (CR) -- (DR);
    \draw[field_arrow] (DR) -- node [right] {$\currentapproxobjectpointcloud$} (ER);

    \draw[field_arrow] (ER) -- (FR);
    \draw[field_arrow] (FR) -- (GR);
    \draw[field_arrow] (GR) -- (HR);
    \draw[field_arrow] (GR) -- (GL);
    \draw[field_arrow] (HR) |- (IL);
    \draw[field_arrow] (IL) -- node [right] {$\currentobjectlist$} (JL);

    \draw[field_arrow] (DL) -- node [right] {$\lastobjectlist$} (EL);
    \draw[model_arrow] (FL) -- (EL);
    \draw[field_arrow] (EL) -- (ER);
    \draw[field_arrow] ([xshift=0.5*\horgap]EL.east) |- (FR);
    \draw[field_arrow] (GL) -- (HL);
    \draw[field_arrow] (HL) -- (IL);
\end{tikzpicture}

%% file: dimension_estimation.tex
\begin{tikzpicture}

    \begin{axis}[ymin=0,
    ymax=3.4,
    xmin=0,
    xmax=5,
    width=7.7cm,
    height=6cm,
    axis line style={white},
    ytick={0.1, 0.5, 0.9},
    yticklabels={$P_{min}$, $P_{ref}$, $P_{max}$},
    xtick={0.0,0.5,2.25,3.0,3.75,5.0},
    xticklabels={0.0,$d^i_j$,$-3\sigma_{\tilde{d}^i}$,$\tilde{d}^i$,$3\sigma_{\tilde{d}^i}$, $d_{max}$},
    xtick pos=bottom
    ]
    \addplot [domain=0:3, samples=101,unbounded coords=jump]{(-2.0*(0.9-0.5)/(1+exp(-1/0.3 * (x-3)))) + 0.9};
    \addplot [domain=3:5, samples=101,unbounded coords=jump]{(-2.0*(0.5-0.1)/(1+exp(-1/0.3 * (x-3)))) + 2.0*0.5 - 0.1};
    \addplot +[mark=none, dashed, black] coordinates {(3, 0) (3, 1)};
    \addplot +[mark=none, dashed, black] coordinates {(3.75, 0) (3.75, 1)};
    \addplot +[mark=none, dashed, black] coordinates {(2.25, 0) (2.25, 1)};
    \addplot +[mark=none, dashed, black] coordinates {(0.5, 0) (0.5, 3.4)};

    \draw[draw=black,<-] (axis cs:0.55,0.85) -- (axis cs:0.75,0.55);
    \node[anchor=west] at (axis cs:0.75,0.55) {$p_d\left(d^i_j \right)$};

    \draw[draw=black] (axis cs:0,0.0) rectangle (axis cs:5.0,1.0);

    \begin{scope}[shift = {(axis cs:0, 3.0)}]
    
        \draw[draw=black] (axis cs:0.55,-0.05) -- (axis cs:3.8,-0.2);
        \draw[draw=black,->] (axis cs:3.8,-0.2) -- (axis cs:3.8,-0.8);

        \draw[fill=gray,draw=black] (axis cs:0,0.0) rectangle (axis cs:0.2,0.2);
        \draw[fill=gray,draw=black] (axis cs:0.2,0.0) rectangle (axis cs:0.4,0.2);
        \draw[fill=gray,draw=black] (axis cs:0.4,0.0) rectangle (axis cs:0.6,0.2);
        \draw[fill=gray,draw=black] (axis cs:0.6,0.0) rectangle (axis cs:0.8,0.2);
        \draw[fill=gray,draw=black] (axis cs:0.8,0.0) rectangle (axis cs:1.0,0.2);
        \draw[fill=gray,draw=black] (axis cs:1.0,0.0) rectangle (axis cs:1.2,0.2);
        \draw[fill=gray,draw=black] (axis cs:1.2,0.0) rectangle (axis cs:1.4,0.2);
        \draw[fill=gray,draw=black] (axis cs:1.4,0.0) rectangle (axis cs:1.6,0.2);
        \draw[fill=gray,draw=black] (axis cs:1.6,0.0) rectangle (axis cs:1.8,0.2);
        \draw[fill=gray!90!white,draw=black] (axis cs:1.8,0.0) rectangle (axis cs:2.0,0.2);
        \draw[fill=gray!80!white,draw=black] (axis cs:2.0,0.0) rectangle (axis cs:2.2,0.2);
        \draw[fill=gray!70!white,draw=black] (axis cs:2.2,0.0) rectangle (axis cs:2.4,0.2);
        \draw[fill=gray!60!white,draw=black] (axis cs:2.4,0.0) rectangle (axis cs:2.6,0.2);
        \draw[fill=gray!50!white,draw=black] (axis cs:2.6,0.0) rectangle (axis cs:2.8,0.2);
        \draw[fill=gray!40!white,draw=black] (axis cs:2.8,0.0) rectangle (axis cs:3.0,0.2);
        \draw[fill=gray!30!white,draw=black] (axis cs:3.0,0.0) rectangle (axis cs:3.2,0.2);
        \draw[fill=gray!20!white,draw=black] (axis cs:3.2,0.0) rectangle (axis cs:3.4,0.2);
        \draw[fill=gray!10!white,draw=black] (axis cs:3.4,0.0) rectangle (axis cs:3.6,0.2);
        \draw[fill=white,draw=black] (axis cs:3.6,0.0) rectangle (axis cs:3.8,0.2);
        \draw[fill=white,draw=black] (axis cs:3.8,0.0) rectangle (axis cs:4.0,0.2);
        \draw[fill=white,draw=black] (axis cs:4.0,0.0) rectangle (axis cs:4.2,0.2);
        \draw[fill=white,draw=black] (axis cs:4.2,0.0) rectangle (axis cs:4.4,0.2);
        \draw[fill=white,draw=black] (axis cs:4.4,0.0) rectangle (axis cs:4.6,0.2);
        \draw[fill=white,draw=black] (axis cs:4.6,0.0) rectangle (axis cs:4.8,0.2);
        \draw[fill=white,draw=black] (axis cs:4.8,0.0) rectangle (axis cs:5.0,0.2);
    \end{scope}
    \begin{scope}[shift = {(axis cs:0, 2.6)}]
        \node[anchor=east] at (axis cs:5.0,0.2) {$t_{k-1}$};
        \draw[draw=black,dashed] (axis cs:0.0,0.0) -- (axis cs:5.0,0.0);
        \node[anchor=east] at (axis cs:5.0,-0.2) {$t_{k}$};
    \end{scope}
    \begin{scope}[shift = {(axis cs:0, 2.0)}]
        \node[anchor=center] at (axis cs:2.5,0.0) {$\mathcal{L}^i_{j,k} = \log \left( \frac{p_d\left(d^i_j\right)}{1-p_d\left(d^i_j\right)} \right) + \alpha \mathcal{L}^i_{j,k-1}$};
    \end{scope}
    \begin{scope}[shift = {(axis cs:0, 1.3)}]
        \draw[draw=black,<-] (axis cs:0.55,0.3) -- (axis cs:0.9,0.55);
        \draw[fill=gray,draw=black] (axis cs:0,0.0) rectangle (axis cs:0.2,0.2);
        \draw[fill=gray,draw=black] (axis cs:0.2,0.0) rectangle (axis cs:0.4,0.2);
        \draw[fill=gray,draw=black] (axis cs:0.4,0.0) rectangle (axis cs:0.6,0.2);
        \draw[fill=gray,draw=black] (axis cs:0.6,0.0) rectangle (axis cs:0.8,0.2);
        \draw[fill=gray,draw=black] (axis cs:0.8,0.0) rectangle (axis cs:1.0,0.2);
        \draw[fill=gray,draw=black] (axis cs:1.0,0.0) rectangle (axis cs:1.2,0.2);
        \draw[fill=gray,draw=black] (axis cs:1.2,0.0) rectangle (axis cs:1.4,0.2);
        \draw[fill=gray,draw=black] (axis cs:1.4,0.0) rectangle (axis cs:1.6,0.2);
        \draw[fill=gray,draw=black] (axis cs:1.6,0.0) rectangle (axis cs:1.8,0.2);
        \draw[fill=gray,draw=black] (axis cs:1.8,0.0) rectangle (axis cs:2.0,0.2);
        \draw[fill=gray,draw=black] (axis cs:2.0,0.0) rectangle (axis cs:2.2,0.2);
        \draw[fill=gray,draw=black] (axis cs:2.2,0.0) rectangle (axis cs:2.4,0.2);
        \draw[fill=gray!85!white,draw=black,draw=black] (axis cs:2.4,0.0) rectangle (axis cs:2.6,0.2);
        \draw[fill=gray!70!white,draw=black,draw=black] (axis cs:2.6,0.0) rectangle (axis cs:2.8,0.2);
        \draw[fill=gray!55!white,draw=black,draw=black] (axis cs:2.8,0.0) rectangle (axis cs:3.0,0.2);
        \draw[fill=gray!45!white,draw=black,draw=black] (axis cs:3.0,0.0) rectangle (axis cs:3.2,0.2);
        \draw[fill=gray!30!white,draw=black,draw=black] (axis cs:3.2,0.0) rectangle (axis cs:3.4,0.2);
        \draw[fill=gray!15!white,draw=black,draw=black] (axis cs:3.4,0.0) rectangle (axis cs:3.6,0.2);
        \draw[fill=white,draw=black] (axis cs:3.6,0.0) rectangle (axis cs:3.8,0.2);
        \draw[fill=white,draw=black] (axis cs:3.8,0.0) rectangle (axis cs:4.0,0.2);
        \draw[fill=white,draw=black] (axis cs:4.0,0.0) rectangle (axis cs:4.2,0.2);
        \draw[fill=white,draw=black] (axis cs:4.2,0.0) rectangle (axis cs:4.4,0.2);
        \draw[fill=white,draw=black] (axis cs:4.4,0.0) rectangle (axis cs:4.6,0.2);
        \draw[fill=white,draw=black] (axis cs:4.6,0.0) rectangle (axis cs:4.8,0.2);
        \draw[fill=white,draw=black] (axis cs:4.8,0.0) rectangle (axis cs:5.0,0.2);
    \end{scope}

    \end{axis}
\end{tikzpicture}

%% file: length_estimation.tex
\begin{tikzpicture}
    \begin{axis}[grid=both,
        width=7.7cm,
        height=4cm,
        reverse legend,
        ymin=-2,
        ymax=12,
        xmin=0,
        xmax=52,
        cycle list name=black white,
        xtick={1,11,21,31,41,51},
        xticklabels={0,1,2,3,4,5},
        x unit={\si{\second}},
        xlabel=Time,
        y unit={\si{\meter}},
        ylabel=Length,
        ticklabel style = {font=\small},
        label style = {font=\small}]

        \draw[draw=black,dashed] (axis cs:0.0,3.0) -- (axis cs:50.0,3.0);

        \addplot[only marks, gray,mark=triangle, error bars/.cd,y dir=both,y explicit] table [x index=0,y index=1,y error index=2,col sep=comma] {length_estimation.csv};
        \addlegendentry{$\tilde{l}$}

        \addplot[name path=lower,forget plot,lightgray] table [x index=0,col sep=comma,y expr=\thisrowno{3} - 3.0*(\thisrowno{4})]{length_estimation.csv};
        
        \addplot[name path=upper,forget plot,lightgray] table [x index=0,col sep=comma,y expr=\thisrowno{3} + 3.0*(\thisrowno{4})]{length_estimation.csv};
        
        \addplot[lightgray] fill between[of=lower and upper];
        \addlegendentry{$\sigma_{\hat{l}}$}

        \addplot[] table [x index=0,y index=3,col sep=comma] {length_estimation.csv};
        \addlegendentry{$\hat{l}$}
    \end{axis}
\end{tikzpicture}   

%% file: length_classification.tex
\begin{tikzpicture}
    \begin{axis}[
        width=7.7cm,
        height=4cm,
        xlabel={Length},
        x unit={\si{\meter}},
        ylabel={Probability Density},
        legend style={legend columns=2},
        grid=both,
        xmin=0, xmax=22,
        ymin=0, ymax=2.1,
        ticklabel style = {font=\small},
        label style = {font=\small},
        colormap name=viridis,
        set layers,
        mark layer=like plot,
        ytick={0,1,2},
        yticklabels={0,1,2},
        xtick={0,5,10,15,20},
        xticklabels={0,5,10,15,20},
    ]
    
    \addplot[gray, domain=0:2, samples=200] {exp(-(x-1.0)^2 / (2*0.2^2)) / (0.2 * sqrt(2*pi))};
    \addlegendentry{Pedestrian}
    
    \addplot[gray, mark=x, mark repeat=10, domain=0.5:3, samples=200] {exp(-(x-1.8)^2 / (2*0.3^2)) / (0.3 * sqrt(2*pi))};
    \addlegendentry{Cyclist}
    
    \addplot[gray, mark=o, mark repeat=10, domain=1:4, samples=200] {exp(-(x-2.2)^2 / (2*0.5^2)) / (0.5 * sqrt(2*pi))};
    \addlegendentry{Motorcycle}
    
    \addplot[gray, mark=triangle, mark repeat=10, domain=2:8, samples=200] {exp(-(x-4.8)^2 / (2*0.74^2)) / (0.74 * sqrt(2*pi))};
    \addlegendentry{Car}
    
    \addplot[gray, mark=diamond, mark repeat=10, domain=4:19, samples=200] {exp(-(x-12.0)^2 / (2*2.0^2)) / (2.0 * sqrt(2*pi))};
    \addlegendentry{Truck}
    
    \end{axis}
\end{tikzpicture}

%% file: classification_estimation.tex
\begin{tikzpicture}
    \begin{axis}[grid=both,
        reverse legend,
        ymin=0,
        ymax=102,
        xmin=0,
        xmax=52,
        height=4cm,
        width=7.7cm,
        cycle list name=black white,
        xtick={1,11,21,31,41,51},
        xticklabels={0,1,2,3,4,5},
        x unit={\si{\second}},
        xlabel=Time,
        ytick={0,25,50,75,100},
        y unit=\text{\%},
        ylabel=Probability,
        ticklabel style = {font=\small},
        label style = {font=\small},
    ]
        \addplot[gray, thick] table [x index=0,y expr=\thisrowno{1}*100,col sep=comma] {classification_estimation.csv};
        
        \addplot[gray, mark=x, mark repeat=2, thick] table [x index=0,y expr=\thisrowno{2}*100,col sep=comma] {classification_estimation.csv};
        
        \addplot[gray, mark=o, mark repeat=2, thick] table [x index=0,y expr=\thisrowno{3}*100,col sep=comma] {classification_estimation.csv};
        
        \addplot[gray, mark=triangle, mark repeat=2, thick] table [x index=0,y expr=\thisrowno{4}*100,col sep=comma] {classification_estimation.csv};
        
        \addplot[gray, mark=diamond, mark repeat=2, thick] table [x index=0,y expr=\thisrowno{5}*100,col sep=comma] {classification_estimation.csv};
    \end{axis}
\end{tikzpicture}   

%% file: existence_histogram.tex
\begin{tikzpicture}
\begin{axis}[
    width=\columnwidth, height=4cm,
    ybar interval,
    xtick=,
    xticklabel=\pgfmathprintnumber\tick--\pgfmathprintnumber\nexttick,
    ymin=0,ymax=20,
    xmin=0,xmax=1,
    xlabel={Mean existance probability},
    ylabel={Number of tracks},
    ticklabel style = {font=\small},
    label style = {font=\small},
    grid=both,
    xtick align=inside,
    width=7.7cm
    ]
    \addplot+ [hist={bins=20,data=x}, fill=lightgray, draw=black] file {existence_histogram.csv};
    \node[fill=white] (A) at (axis cs:0.2, 7) {False positives};
    \draw[->] (A) -- (axis cs:0.28, 3);

    \node[fill=white, anchor=west] (B) at (axis cs:0.4, 18) {True positives};
    \draw[dashed] (axis cs:0.4, 0) -- (axis cs:0.4, 20);
    \draw[->] (axis cs:0.4, 16) -- (axis cs:0.65, 16);

\end{axis}
\end{tikzpicture}

%% file: error_tracking.tex
\begin{tikzpicture}
	\begin{axis} [
	at={(0cm,0cm)},
	ylabel={Deviation}, 
    y unit={\si{\meter}},
	xtick=data,
	enlarge x limits=0.5,
	width=5.9cm,
    height=4.0cm,
	xticklabels={Forward,Lateral},
	axis y line*=left,
	ymin=-3,
	ymax=3,
	ymajorgrids
	]
		\addplot [box plot median] table {error_tracking.csv};
		\addplot [box plot box] table {error_tracking.csv};
		\addplot [box plot top whisker] table {error_tracking.csv};
		\addplot [box plot bottom whisker] table {error_tracking.csv};
	\end{axis}

	\begin{axis} [
	at={(4.3cm,0cm)},
	ylabel={Deviation}, 
    y unit={\si{\degree}},
	xtick=data,
	enlarge x limits=0.5,
	width=3.4cm,
    height=4.0cm,
	xticklabels={Orientation},
	ylabel near ticks, yticklabel pos=right,
	ymin=-30,
	ymax=30,
	xmin=-1,
	xmax=1,
	ymajorgrids
	]
		\addplot [box plot median] table {error_tracking_2.csv};
		\addplot [box plot box] table {error_tracking_2.csv};
		\addplot [box plot top whisker] table {error_tracking_2.csv};
		\addplot [box plot bottom whisker] table {error_tracking_2.csv};
	\end{axis}
\end{tikzpicture}

%% file: error_dimension.tex
\begin{tikzpicture}
	\begin{axis} [
	at={(0cm,0cm)},
	ylabel={Deviation}, 
    y unit={\si{\meter}},
	xtick=data,
	enlarge x limits=0.5,
	width=7.7cm,
    height=4.0cm,
	xticklabels={Length, Width, Height},
	ymin=-2,
	ymax=2,
	ymajorgrids
	]
		\addplot [box plot median] table {error_dimension.csv};
		\addplot [box plot box] table {error_dimension.csv};
		\addplot [box plot top whisker] table {error_dimension.csv};
		\addplot [box plot bottom whisker] table {error_dimension.csv};
	\end{axis}
\end{tikzpicture}

%% file: timing_histogram.tex
\begin{tikzpicture}
\begin{axis}[
    width=\columnwidth, height=4cm,
    ybar interval,
    xtick={0,20,40,60,80,100,140},
    xticklabel=\pgfmathprintnumber\tick--\pgfmathprintnumber\nexttick,
    ymin=0,ymax=3500,
    xmin=0,xmax=140,
    xlabel={Total processing runtime},
    x unit={ms},
    ylabel={Number of frames},
    ticklabel style = {font=\small},
    label style = {font=\small},
    grid=both,
    xtick align=inside,
    width=7.7cm
    ]
    \addplot+ [hist={bins=20,data=x}, fill=lightgray, draw=black] file {timing_histogram.csv};

    \node[fill=white, anchor=west] (B) at (axis cs:105, 3000) {Deadline};
    \draw[dashed] (axis cs:100, 0) -- (axis cs:100, 3500);
    \draw[->] (axis cs:130, 2500) -- (axis cs:100, 2500);

    \node[fill=white, anchor=south] (B) at (axis cs:120, 550) {$n=11$};
    \draw[->] (axis cs:120, 500) -- (axis cs:120, 100);

\end{axis}
\end{tikzpicture}